\documentclass[10pt,twocolumn,letterpaper]{article}

\usepackage[pagenumbers]{cvpr} % To force page numbers, e.g. for an arXiv version

\usepackage{amsmath,amsfonts}
\usepackage{algorithmic}
\usepackage{algorithm}
\usepackage{array}
\usepackage{textcomp}
\usepackage{url}
\usepackage{verbatim}
\usepackage{graphicx}
\usepackage{subcaption}
\usepackage[export]{adjustbox}
\usepackage{amssymb}

\newlength{\tempdim}
\newcommand{\authorskip}{\hspace{8mm}}

\usepackage[dvipsnames]{xcolor}

\definecolor{cvprblue}{rgb}{0.21,0.49,0.74}
\usepackage[pagebackref,breaklinks,colorlinks,citecolor=cvprblue]{hyperref}

\title{Highly Detailed and Temporal Consistent Video Stylization via \\Synchronized Multi-Frame Diffusion}

\author{
    Minshan Xie$^{1}$ \authorskip Hanyuan Liu$^{2}$ \authorskip Chengze Li$^{3}$ \authorskip Tien-Tsin Wong$^{1}$\\
  $^{1}$The Chinese University of Hong Kong \quad $^{2}$City University of Hong Kong \\
  $^{3}$Caritas Institute of Higher Education\\
  {\tt\small \{msxie,ttwong\}@cse.cuhk.edu.hk, hy.liu@cityu.edu.hk, czli@cihe.edu.hk}
}

\begin{document}
\maketitle
\begin{abstract}

Text-guided video-to-video stylization transforms the visual appearance of a source video to a different appearance guided on textual prompts. 
Existing text-guided image diffusion models can be extended for stylized video synthesis. However, they struggle to generate videos with both highly detailed appearance and temporal consistency. 
In this paper, we propose a synchronized multi-frame diffusion framework to maintain both the visual details and the temporal consistency. Frames are denoised in a synchronous fashion, and more importantly, information of different frames is shared since the beginning of the denoising process. Such information sharing ensures that a consensus, in terms of the overall structure and color distribution, among frames can be reached in the early stage of the denoising process before it is too late. 
The optical flow from the original video serves as the connection, and hence the venue for information sharing, among frames. 
We demonstrate the effectiveness of our method in generating high-quality and diverse results in extensive experiments. Our method shows superior qualitative and quantitative results compared to state-of-the-art video editing methods.

\end{abstract}    
\section{Introduction}
\label{sec:intro}

% Background and motivation
Video-to-video stylization or conversion, takes a source video (e.g. live action video) as input and converts it to a target one with the desired visual effects (e.g. cartoon style, or photorealistic one with the change of person's identity/hairstyle/dressing, etc). It can be regarded as a generalized rotoscoping, not only to produce cartoon animation, but more general ones. Due to its convenience and generality, there has be a large demand in the video content production, as observed in social platforms, such as YouTube and TikTok, even though the produced videos exhibit significant visual and temporal inconsistencies.

\begin{figure}[t!]
    \centering
    \includegraphics[width=\linewidth]{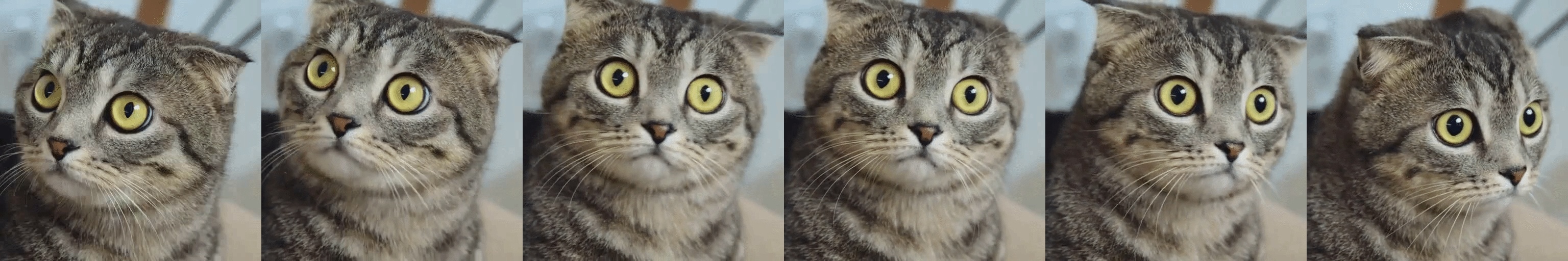}\\
    \includegraphics[width=\linewidth]{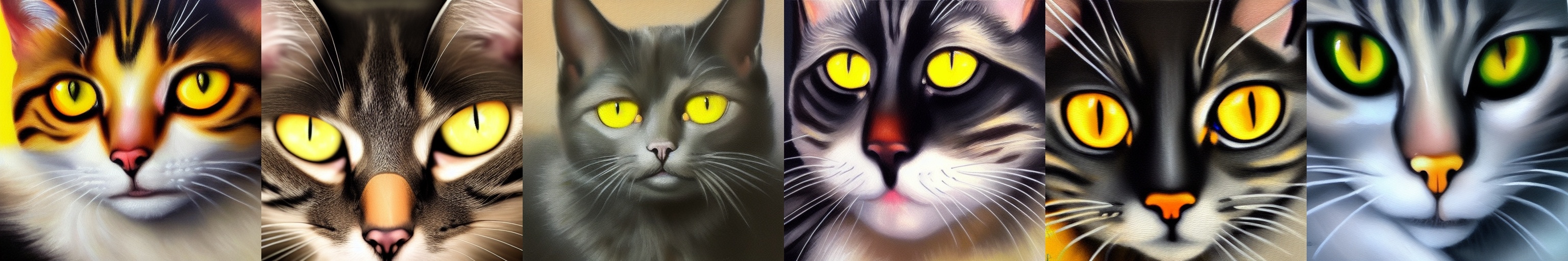}\\
    \includegraphics[width=\linewidth]{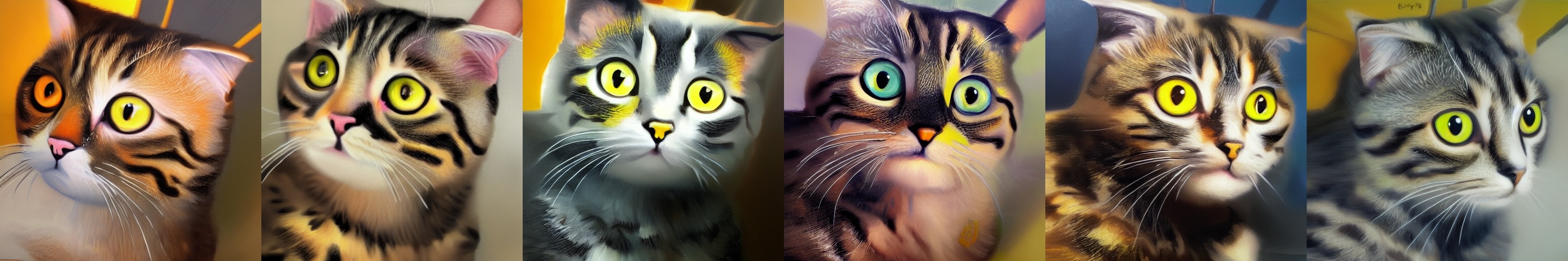}\\
    \includegraphics[width=\linewidth]{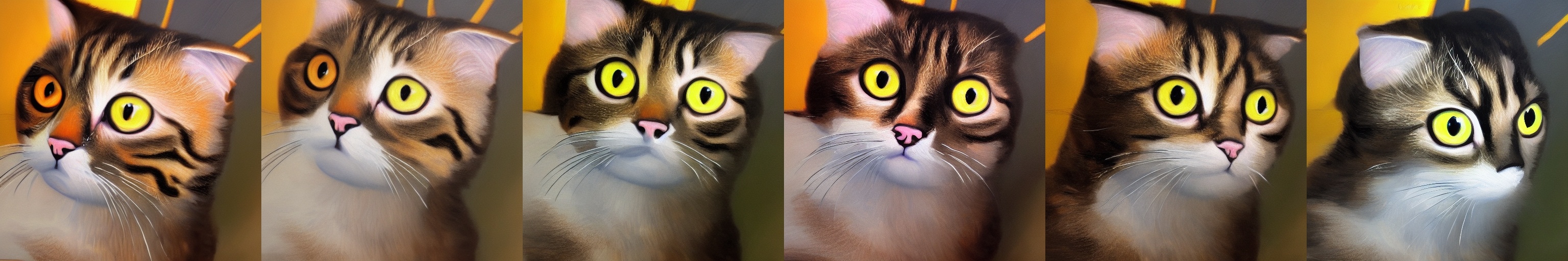}\\
    \includegraphics[width=\linewidth]{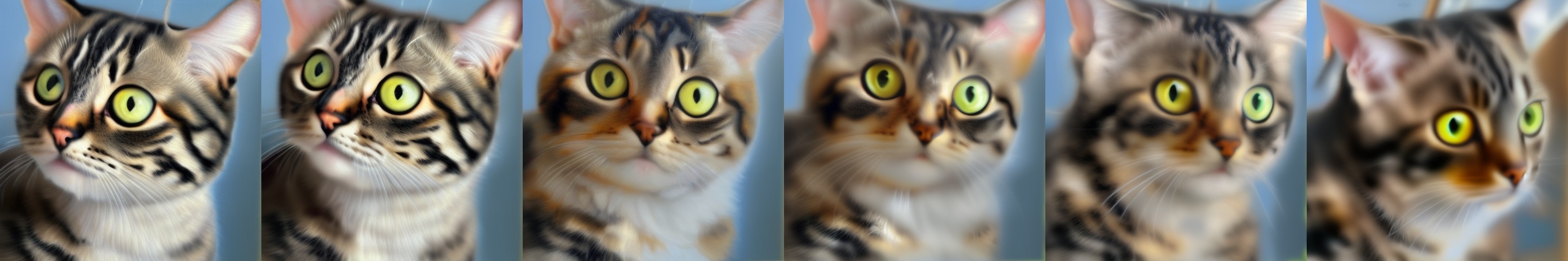}\\
    \includegraphics[width=\linewidth]{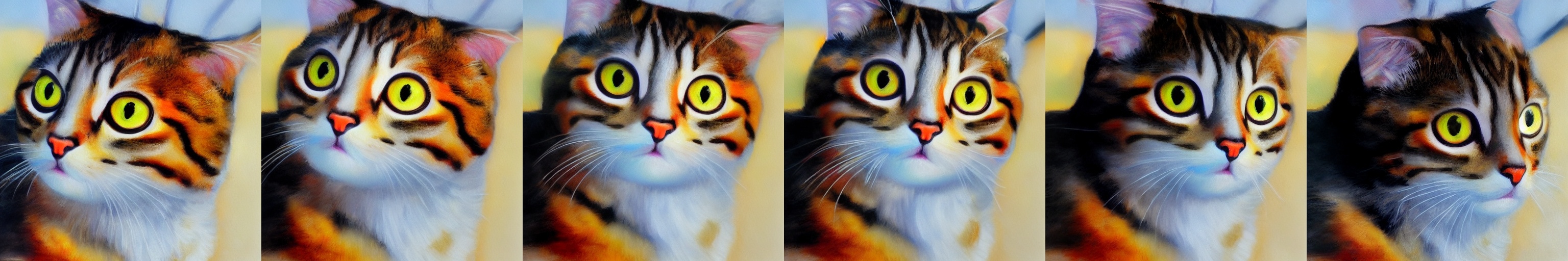}    
    \caption{Our method can generate stylized frames with local visual consistency. From top to bottom: original video, SD~\cite{rombach2022high}, ControlNet~\cite{zhang2023adding}, Text2Video-Zero~\cite{khachatryan2023text2video}, Rerender-A-Video~\cite{yang2023rerender} and ours. Text prompt: {\footnotesize\tt "A cat with yellow eyes, oil painting."} Readers are encouraged to zoom in to better compare the fine details from different methods.}
    \label{fig:tempinconsist}
\end{figure}

With the advances of large-scale data trained diffusion models, text-to-image (T2I) diffusion models ~\cite{ramesh2022hierarchical,rombach2022high,ho2022imagen} present the exceptional ability in generating diverse and high-quality images, and more importantly, its conformity to the text description given by users. 
Subsequent works based on T2I models~\cite{meng2021sdedit,gafni2022make,hertz2022prompt,brooks2023instructpix2pix,kawar2023imagic} further demonstrate its image editing functionality. 
Therefore, it is natural to apply these  T2I methods to the above video stylization task
~\cite{khachatryan2023text2video,yang2023rerender,chai2023stablevideo} by applying the {\em pretrained} T2I diffusion model on each frame individually (the second row of Fig.~\ref{fig:tempinconsist}). However, even with per-frame constraints from the ControlNet~\cite{zhang2023adding}, the direct T2I  application  cannot maintain the temporal consistency and leads to severe flickering artifacts (the third row of Fig.~\ref{fig:tempinconsist}).

To maintain the temporal consistency, one can apply text-to-video (T2V) diffusion models~\cite{ho2022imagen,molad2023dreamix,guo2023animatediff}, but with a trade-off of high computational training cost. This may not be cost effective. Some zero-shot methods~\cite{khachatryan2023text2video,cong2023flatten} imposes cross-frame constraints on the latent features for temporal consistency, but these constraints are limited to global styles and are unable to preserve low-level consistency, which may still exhibit flickering local structures (the fourth row of Fig.~\ref{fig:tempinconsist}). 
A few methods utilize the optical flow to improve the low-level temporal consistency of the resultant videos. They typically warp from one frame to another, using the optical flow,  patch the unknown region ~\cite{yang2023rerender,chai2023stablevideo}, and followed by a post-processing smoothing~\cite{yang2023rerender,khachatryan2023text2video} for consistent appearance (warp-and-patch approach), which inevitably leads to alignment artifacts or over-blurriness (the fifth row of Fig.~\ref{fig:tempinconsist}). 
It remains challenging to simultaneously achieve the highly detailed fidelity, the conformity to text prompt, and the temporal consistency throughout the entire video sequence.

In this paper, instead of using the optical flow for warp-and-patch, we utilize the correspondence sites, determined from the optical flow, as {\em portals for information sharing} among the frames. 
Such information sharing among frames is performed between each denoising step, hence we called it {\em synchronized multi-frame diffusion}.
It is crucial for originally separated diffusion processes of frames to reach a {\em consensus}, in terms of overall visual layout and color distribution, in the early stage of the diffusion process, before it is too late to fix. 
To achieve this, we design a multi-frame fusion stage on top of the existing diffusion model, which adds temporal consistency constraints to the intermediate video frames generated at each diffusion step. The visual content is unified among frames through consensus-based information sharing. We first propagate the content of each frame to overlapping regions in other frames. Then, each frame is updated (denoised) by fusing the propagated (shared) information received from all other frames.
However, we observed that global-scale and medium-scale structure consensus can be achieved in  the early denoising steps, but fine-scale detail consensus fails to be achieved with the misaligned detail generated at the later denoising steps. To prevent the generated details from being smoothed out, we propose an alternative propagation strategy that propagates the details of randomly selected frames to overwrite the overlapping regions in other frames. As each frame has an equal opportunity to propagate the details, a pseudo-equal sharing way is achieved.

We conduct extensive qualitative and quantitative experiments to demonstrate the effectiveness of our method. Our method achieves outstanding performance compared with state-of-the-art methods in all evaluated metrics. It strikes a nice balance in terms of temporal consistency and semantic conformity to user prompts.
Our contributions are summarized as follows:
\begin{itemize}
    \item Instead of warp-and-patch approach, our zero-shot method is designed based on a consensus approach, in which all frames contribute to the generation of stylized content, in an equal and synchronized fashion. 
    \item We propose to seamlessly blend the shared content from different frames using a novel Multi-Frame Fusion. 
\end{itemize}
\section{Related Work}
\label{sec:relatedwork}

\paragraph{Text-Driven Image Editing.}

Advancements in computer vision have led to significant progress in natural image editing. Before the rise of diffusion models~\cite{song2020denoising,ho2020denoising}, various GAN-based approaches~\cite{wang2018high,park2019semantic,goodfellow2020generative,gal2022stylegan} achieved commendable results. 
The emergence of diffusion models has elevated the quality and diversity of edited content even further. SDEdit~\cite{meng2021sdedit} introduces noise and corruptions to an input image and then leverages diffusion models to reverse the process, enabling effective image editing. But, it suffers from the loss of fidelity. 
Prompt-to-Prompt~\cite{hertz2022prompt} and Plug-and-Play~\cite{tumanyan2023plug} perform semantic editing by blending activations from original and target text prompts.
UniTune~\cite{valevski2023unitune} and Imagic~\cite{kawar2023imagic} focus on finetuning a single image for improved editability while maintaining fidelity. 
Researchers have also explored aspects like controllability~\cite{saharia2022palette,li2023gligen,zhang2023adding,cao2023image,huang2023composer} and personalization~\cite{ruiz2023dreambooth,gal2022image} in diffusion-based generation, enhancing our understanding of how to tailor diffusion models to specific editing needs. 
Our proposed method builds upon existing image editing techniques~\cite{zhang2023adding,mou2023t2i,parmar2023zero} to preserve structural integrity and generate videos with temporal consistency.

\paragraph{Text-Driven Video Editing.}
Video editing poses unique challenges for diffusion-based methods compared to image editing, primarily due to the intricate requirements of geometric and temporal consistency. While image editing has seen significant progress, extending these advancements to videos remains a complex task. 
Text-to-Video (T2V) Diffusion Models~\cite{ho2022video} have emerged as a promising avenue. These models build upon the 2D U-Net architecture used in image models but extend it to a factorized space-time UNet~\cite{ho2022imagen,singer2022make,yu2023video,zhou2022magicvideo,guo2023animatediff}. Dreamix~\cite{molad2023dreamix} focuses on motion editing by developing a text-to-video backbone while ensuring temporal consistency. Make-A-Video~\cite{singer2022make} leverages unsupervised video data and learns movement patterns to drive the image model. However, these methods require substantial video data for training.
StableVideo~\cite{chai2023stablevideo} employs a compressed representation as the propagator for consistent video editing. It generates the appearance of the next frame based on warped information from the previous one. However, it requires additional training for the compressed representation and may involve suboptimal training for an aggregation network to unify the edited foreground appearance.

Some recent efforts aim to make video editing more cost-effectively. Methods like Tune-a-Video~\cite{wu2023tune} , UniTune~\cite{valevski2023unitune}, and Imagic~\cite{kawar2023imagic} propose fine-tuning pre-trained T2I diffusion models on single videos to achieve consistent video editing. However, modeling complex motion remains a challenge. 
Some zero-shot methods~\cite{khachatryan2023text2video}, such as Text2Video-Zero~\cite{khachatryan2023text2video} and ControlVideo~\cite{zhang2023controlvideo}, impose cross-frame constraints on latent features for temporal consistency and use ControlNet~\cite{zhang2023adding} for controllable video editing. However, these constraints are often limited to global styles and struggle to preserve low-level visual consistency.

Several methods have emerged to address the challenge of maintaining consistency across frames while preserving visual quality, relying on key frames~\cite{jamrivska2019stylizing,texler2020interactive,xu2022temporally} or optical flow~\cite{ruder2016artistic} to propagate contents between frames. 
FLATTEN~\cite{cong2023flatten} introduces a flow-guided attention mechanism that leverages optical flow to guide the attention module during the diffusion process. However, as these methods operate in the latent domain, they may lead to low-level visual inconsistencies.
Rerender-A-Video~\cite{yang2023rerender} utilizes optical flow to apply dense cross-frame constraints. It gradually inpaints the next frame by warping the overlapping region from the previous one. The fused regions combine to form the final output. However, the results tend to be blurry, as a smoothing operation is employed to avoid artifacts during fusion. Additionally, it may introduce inconsistent styles for disoccluded regions.
Different with existing methods which follow a warp-and-patch strategy and a subsequent merging step, we propose to impose the temporal coherence with synchronized multi-frame diffusion to reach a consensus for all frames, in which all frames contribute more-or-less equally.

\section{Preliminary}

\textbf{Diffusion Models}~\cite{sohl2015deep} are powerful probabilistic models that gradually denoise data, effectively learning the reverse process of a fixed Markov Chain~\cite{ho2020denoising,dhariwal2021diffusion}. These models aim to learn the underlying data distribution $p(x_0)$ by iteratively denoising a normally distributed variable. The denoising process involves a sequence of denoising networks, denoted as $\epsilon_\theta(x_{t},t);\, t=1,\dots, T$. The model is trained to predict a denoised variant of its input $x_{t-1}$ from $x_t$, where $x_{t-1}$ and $x_t$ represents the noisy version of the original input $x_0$. Besides, the problem can also be transformed to predict a clean version $x_{0|t}$ from $x_t$ as we can sample $x_{t-1}$ based on $x_{0|t}$ with a deterministic DDIM sampling~\cite{song2020denoising,song2020score}.

\noindent \textbf{Latent Diffusion Models (LDMs)}~\cite{rombach2022high} employ perceptual compression through an autoencoder architecture, consisting of an encoder $\mathcal{E}$ and a decoder $\mathcal{D}$. LDMs learn the conditional distribution $p(z|y)$ of condition $y$, where $z$ represents the latent representation obtained from the encoder $\mathcal{E}$. The decoder $\mathcal{D}$ aims to reconstruct the original input $x$ from this latent representation, i.e., $\mathcal{E}(x)=z$, $\mathcal{D}(\mathcal{E}(x))\approx x$. The loss function quantifies the discrepancy between the noisy input and the output of the neural backbone. 
The neural backbone is generally realized as a denoising U-Net with cross-attention conditioning mechanisms~\cite{vaswani2017attention} to accept additional conditions.

\noindent \textbf{Conditional Generation.} Natural language is flexible for global style editing but has limited spatial control over the output (the second row in Fig.~\ref{fig:tempinconsist}). To improve spatial controllability, Zhang et al.~\cite{zhang2023adding} introduced a side path called {\em ControlNet} for Stable Diffusion to accept extra conditions, such as edges, depth, and human pose. ControlNet is often used to provide structure guidance from the input video to improve temporal consistency. However, ControlNet alone is insufficient to ensure  medium- and fine-scale consistencies in terms of color and texture, across the frames (the third row in Fig.~\ref{fig:tempinconsist}).
To address this issue, cross-frame attention mechanisms~\cite{khachatryan2023text2video} are further applied to all sampling steps for global style consistency on the latent features. These constraints are limited to global styles and lead to color jittering and fine-scale visual inconsistencies (the forth row in Fig.~\ref{fig:tempinconsist}). 

In contrast, we aim to generate a new video, in a style specified by text prompt, not just with temporal consistency, but also visual consistency in global, medium and fine scales. These consistencies are accomplished via sharing information among frames, using our proposed Synchronized Multi-Frame Diffusion process. 

\section{Synchronized Multi-Frame Diffusion}
\label{sec:method}

\begin{figure*}[t!]
    \centering
    \includegraphics[width=\linewidth]{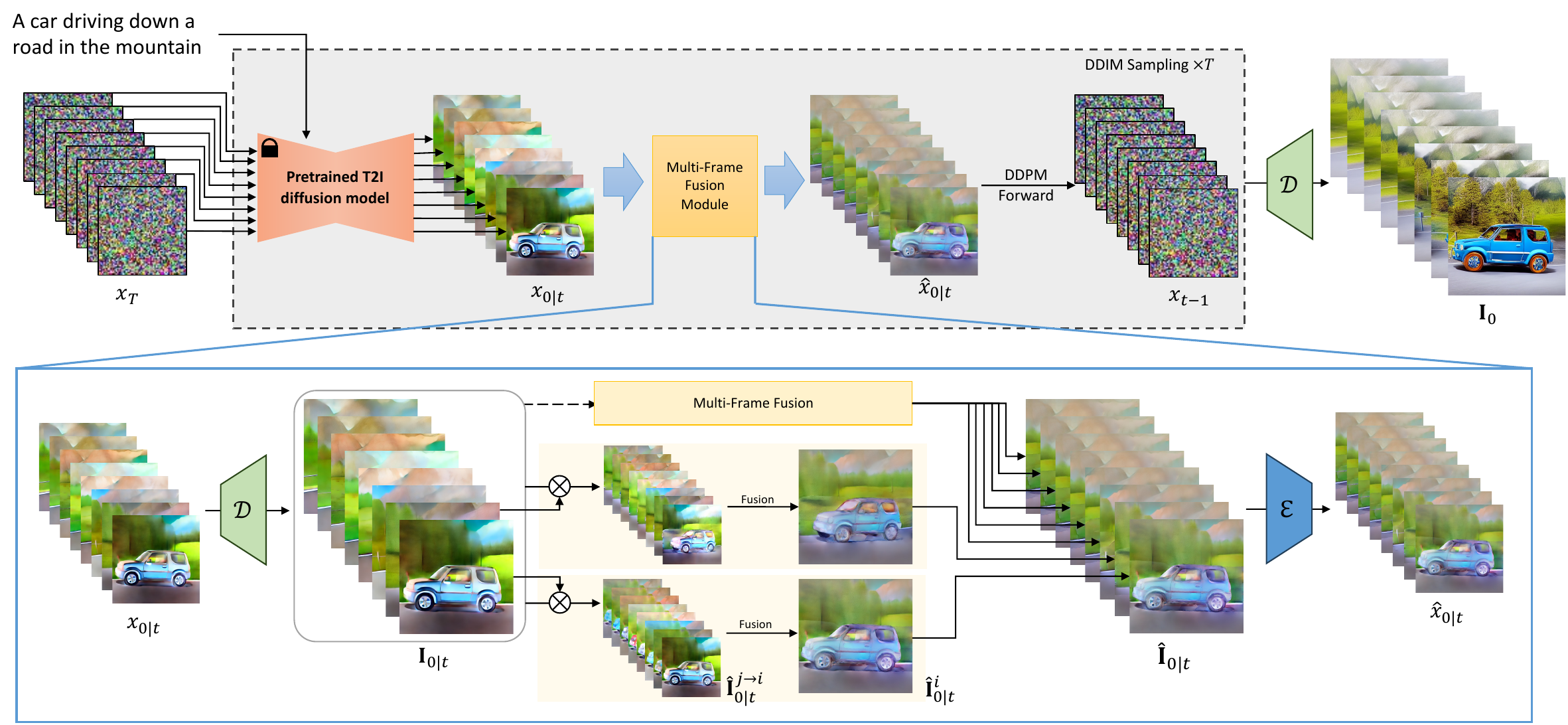}
    \caption{Framework of the proposed zero-shot text-guided video stylization. We first adopt a pretrained T2I model with cross-frame attention layers to generate stylized frames with global style consistency. The stylized frames are refined to render consistent frames in terms of visual content, color distribution, and temporal motion, using our Multi-Frame Fusion Module at each denoising step. }
    \label{fig:overview}
\end{figure*}

Given a video with $N$ frames $\{\textbf{I}_i\}^N_{i=0}$, our goal is to render it into a new video $\{\textbf{I}'_i\}^N_{i=0}$ in a style specified by a text prompt.
The stylized video shall mimic the motion of the original video, and maintain the temporal consistency and visual consistencies in all scales. 
To achieve this, we first assign a T2I diffusion process to each frame to generate the desired style. The major challenge here is on how to generate consistent frames in all visual scales. Instead of warping the generated content from one view to another and then smoothing as in previous approaches~\cite{yang2023rerender,chai2023stablevideo}, we propose a consensus-based approach in which all frames share their latent information among each other during each denoising time step. We call this method, {\em Synchronized Multi-Frame Diffusion} (SMFD).

As a frame must overlap with its neighboring frames, the generated content within the overlapping regions should be consistent. In other words, these overlapping regions (obtained via optical flow) can serve as a venue for latent information sharing among the frame diffusion processes.  For each denoising time-step, the latent information from all frame diffusion processes are first combined before  the next round of denoising. 
Fig.~\ref{fig:overview} shows our proposed video stylization framework. 

To combine the contribution from all overlapped neighboring frames, we can warp the content from all involved neighboring frames to the current frame of interest and fuse them together using a Poisson solver~\cite{perez2003poisson,sun2004poisson}. Disoccluded regions and image border  in the warped content can be seamlessly handled in the gradient domain during the Poisson solving. Such fusion is performed  for each frame with diffusion attached. This Multi-Frame Fusion Module is detailed in Sec.~\ref{subsec:mffm}. With this information sharing among frames, the consensus in terms of color distribution and the overall structure can be reached in the early stage of the denoising process.

Although directly combining content from all involved frames can well unify the coarse-level visual content among frames during the early denoising steps ({\em semantic reconstruction stage}), it smoothens out the high-frequency details in the later denoising steps ({\em detail refinement stage}), leading to over-blurriness. 
To avoid smoothing out the fine details, we adopt an alternating propagating strategy during the detail refinement stage. We propagate the generated details of a randomly selected frame to overlapping region in other frames and overwrite (instead of fusing) the conflict details. A random frame is selected in each denoising step to encourage the contribution from involving frames.
With such design, we can achieve both highly detailed fidelity and temporal consistency throughout the entire video sequence. 
In all our experiments, we treat the first half of denoising steps, $\frac{T}{2}<t<T$, as the semantic reconstruction stage, and the second half, $0<t\leq \frac{T}{2}$, as the detail refinement stage.

\subsection{Multi-Frame Fusion Module}
\label{subsec:mffm}

\begin{figure}[t!]
    \centering
    \begin{subfigure}{0.24\linewidth}
        \includegraphics[width=\linewidth]{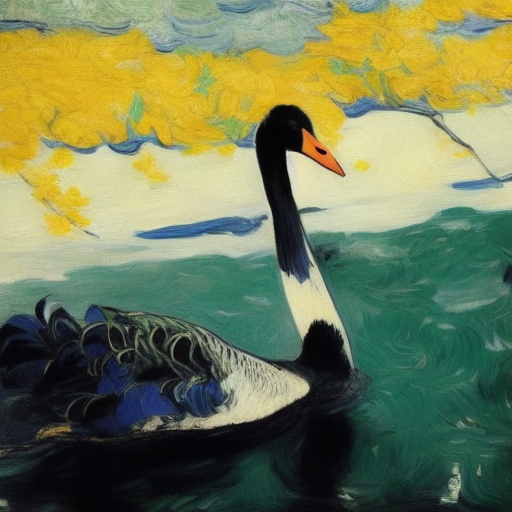}
        \caption{}
    \end{subfigure}
    \begin{subfigure}{0.24\linewidth}
        \includegraphics[width=\linewidth]{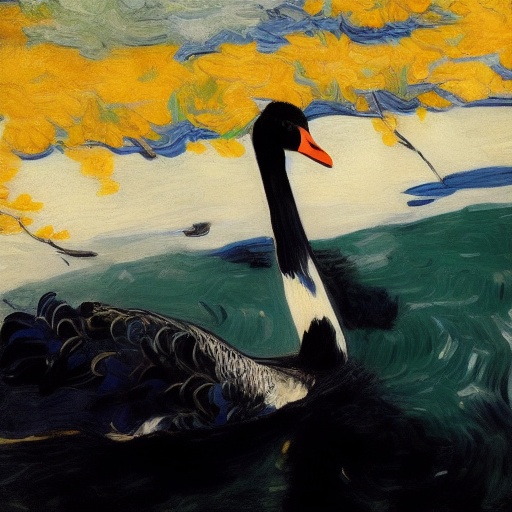}
        \caption{}
    \end{subfigure}
    \begin{subfigure}{0.24\linewidth}
        \includegraphics[width=\linewidth]{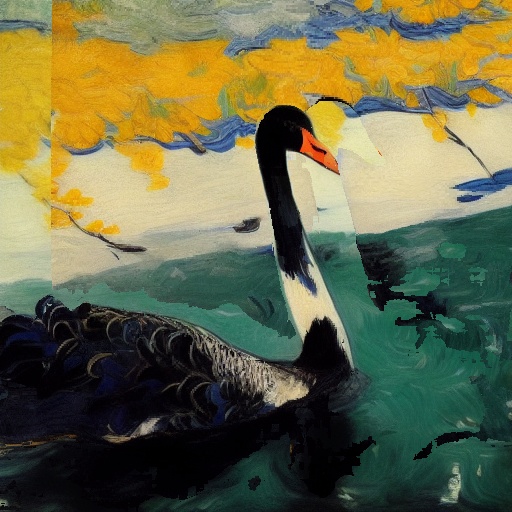}
        \caption{}
    \end{subfigure}
    \begin{subfigure}{0.24\linewidth}
        \includegraphics[width=\linewidth]{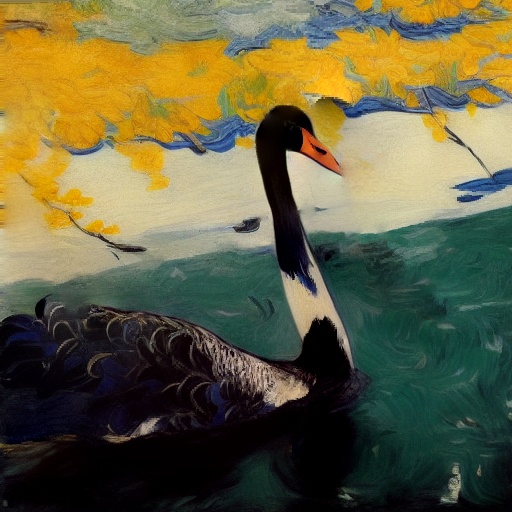}
        \caption{}
    \end{subfigure}
    \caption{We use poisson image editing to seamlessly blend the overlapping region. (a) $I^i_t$, (b) $I^j_t$, (c) Copy-and-paste exhibits obvious seams, (d) Poisson blending.}
    \label{fig:seamless}
\end{figure}

In our framework, we adopt the pretrained T2I diffusion models with structure control~\cite{zhang2023adding} and cross-frame attention mechanism~\cite{khachatryan2023text2video} to create stylized frames $\{\textbf{I}_i^t\}^N_{i=0}$. 
In order to achieve pixel-level visual consistency, we perform multi-frame fusion in the image domain.
We tackle the problem by updating each frame with the appearance information received from other frames, thereby achieving consensus among all frames. One important question is how to propagate the information of appearance across frames to achieve consistency. A simple way is to directly update the current frame using the overlapping region of other frames. However, it is obvious that there will be seams between the updated overlapping region and the rest of the region (Fig.~\ref{fig:seamless}(c)), due to the disocclusion. 

\begin{figure}[t!]
    \centering
    \includegraphics[width=\linewidth]{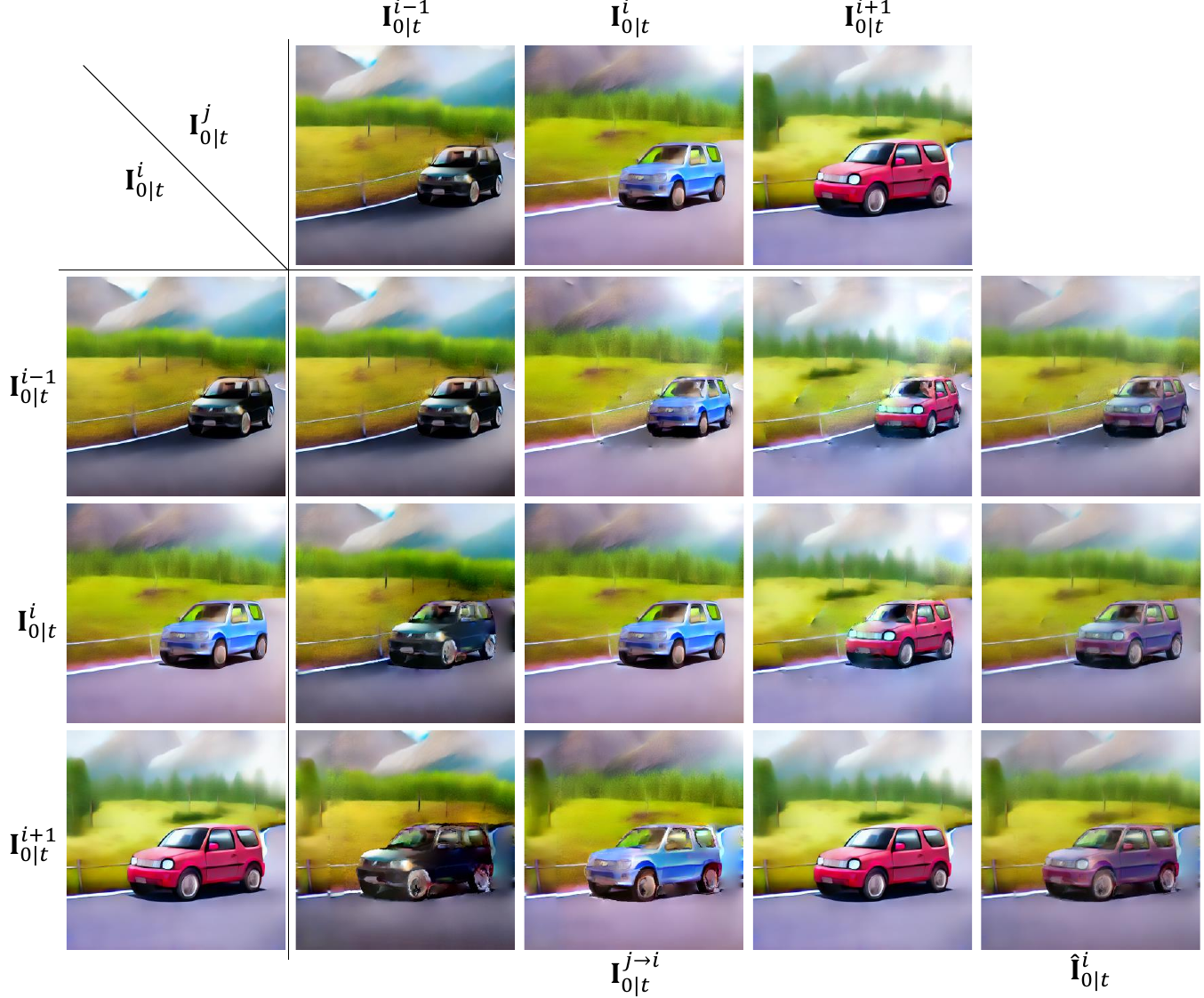}
    \caption{For each frame, we can generate multiple candidates following similar color distribution with the other frames. Thus, the fused frames can have similar appearance among all frames.}
    \label{fig:mffmcase}
\end{figure}

Inspired by Ebsynth~\cite{jamrivska2019stylizing}, we propose to blend the warped appearance from other frames in the gradient domain, and then solve for the images using Poisson equation. This generates multiple seamless candidates. These seamless candidates can further update the current frame without producing obvious seams. 
For every frame $\hat{\textbf{I}}^j_{0|t}$, we can warp it to the pose of frame $\hat{\textbf{I}}^i_{0|t}$, and yield a candidate image $\hat{\textbf{I}}^{j\rightarrow i}_{0|t}$, in which its  appearance follows $\hat{\textbf{I}}^j_{0|t}$, but pose follows $\hat{\textbf{I}}^i_{0|t}$. Fig.~\ref{fig:mffmcase} shows all candidate images of a 3-frame video. Each of the black, blue and red cars are warped to all possible poses (Fig.~\ref{fig:mffmcase}, middle 3$\times3$ table). By combining all candidates, the fused frame can have similar appearance (car with a mixture appearance of black, blue and red) among all frames (Fig.~\ref{fig:mffmcase}, the rightmost column), and thereby achieving the visual consistency.

\begin{figure}[t!]
    \centering
    \begin{subfigure}{0.19\linewidth}
        \includegraphics[width=\linewidth]{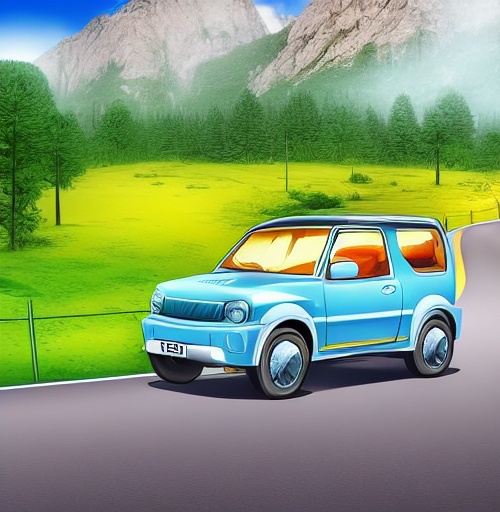}\\
        \includegraphics[width=\linewidth]{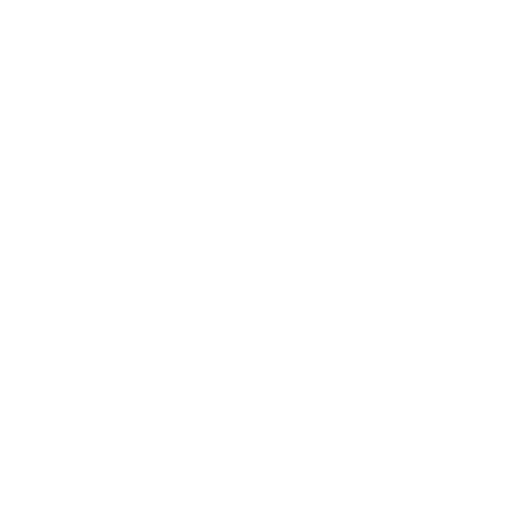}
        \caption{}
    \end{subfigure}
    \begin{subfigure}{0.19\linewidth}
        \includegraphics[width=\linewidth]{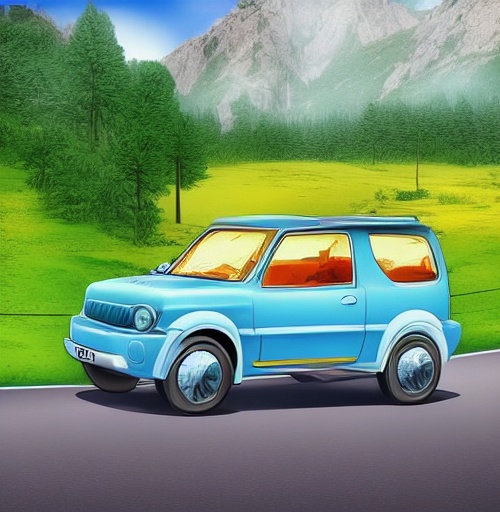}\\
        \includegraphics[width=\linewidth]{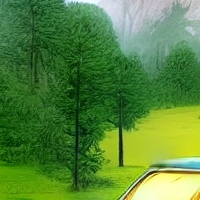}
        \caption{}
    \end{subfigure}
    \begin{subfigure}{0.19\linewidth}
        \includegraphics[width=\linewidth]{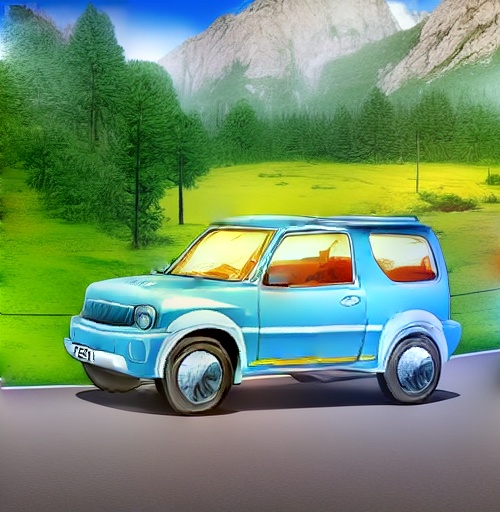}\\
        \includegraphics[width=\linewidth]{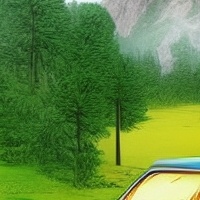}
        \caption{}
    \end{subfigure}
    \begin{subfigure}{0.19\linewidth}
        \includegraphics[width=\linewidth]{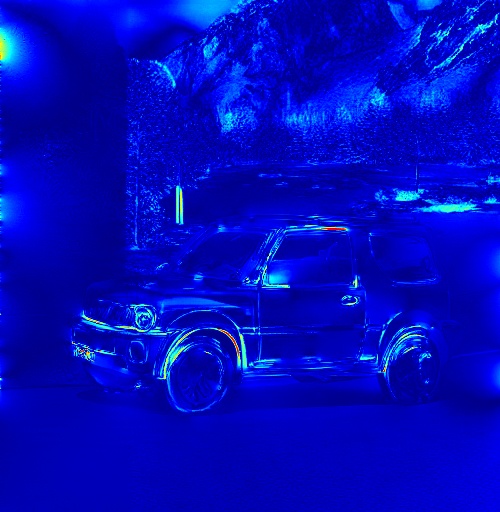}\\
        \includegraphics[width=\linewidth]{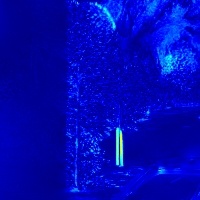}
        \caption{}
    \end{subfigure}
    \begin{subfigure}{0.19\linewidth}
        \includegraphics[width=\linewidth]{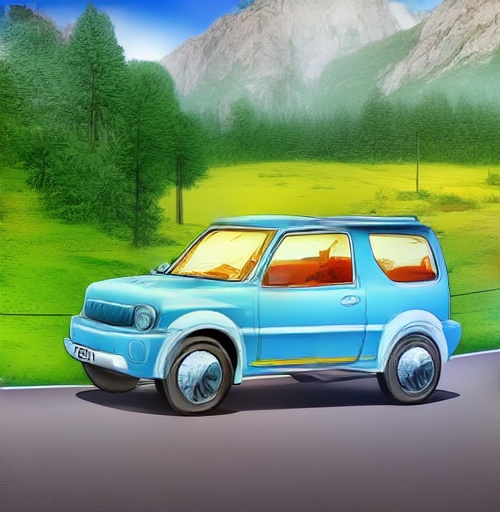}\\
        \includegraphics[width=\linewidth]{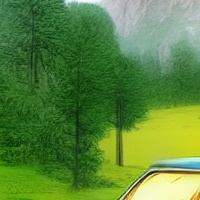}
        \caption{}
    \end{subfigure}
    \caption{Details in different frames with misalignment can lead to blurriness after averaging. (a) Frame 1, (b) Frame 2, (c) Poisson blended image, (d) difference of (b)\&(c), (e) fused image of (b)\&(c). }
    \label{fig:misalign}
\end{figure}

While the overall semantic structure and color distribution can be preserved by above fusion, the details may be damaged due to misalignment of fine textures from different frames (Fig.~\ref{fig:misalign}). To generate consistent frames with high fidelity, we adopt a pseudo-equal sharing way by alternatively propagating the details of randomly selected frames to overwrite the conflict textures during the later denoising steps.

\paragraph{Shared information propagation.}
Each predicted frame $\textbf{I}^i_{0|t}$ is firstly warped to other frames using optical flow and generates candidate edited frames for combination. 
However, directly copying the overlapped region from other frames and pasting it onto the current frame leads to large abrupt intensity changes or seams. Thus, we propose to seamlessly blend the occluded regions to the warped frame using a Poisson solver~\cite{perez2003poisson,sun2004poisson}. The idea is to reconstruct pixels in the blending region such that the  boundary of warped content owns a zero gradient. 
Fig.~\ref{fig:seamless}(c) shows the obvious seam of the warped image boundary if we simply copy-and-paste the warped content, while no seam is observed if we adopt the Poisson blending in Fig.~\ref{fig:seamless}(d). 
Then we can generate a candidate (warped) frames $\hat{\textbf{I}}^i_{0|t}$ at timestep $t$ with 
\begin{equation}
    \hat{\textbf{I}}^{j\rightarrow i}_{0|t} = {\rm PIE}(\textbf{I}^i_{0|t}, w_j^i(\textbf{I}^j_{0|t}), M_j^i),
\end{equation}
where $w_j^i$ and $M_j^i$ denote the optical flow and occlusion mask from $\textbf{I}_j$ to $\textbf{I}_i$, respectively. ${\rm PIE}(\cdot,\cdot,\cdot)$ donates the Poisson solver~\cite{perez2003poisson,sun2004poisson} which seamlessly blends the masked region of $\textbf{I}^i_{0|t}$ into $w_j^i(\textbf{I}^j_{0|t})$. Thus, $\hat{\textbf{I}}^{j\rightarrow i}_{0|t}$ can follow the color appearance of $w_j^i(\textbf{I}^j_{0|t})$.

\paragraph{Candidates Fusion at Semantic Reconstruction Stage.}
We then need to fuse these candidate frames to guarantee consistent geometric and appearance among all stylized frames. For frame $\textbf{I}^i_{0|t}$, we can obtain $N-1$ candidate frames $\hat{\textbf{I}}^{j\rightarrow i}_{0|t}$ which has the same geometric structure but different color appearances. 
The updated frames is the simply average value of all candidate frames. 
\begin{equation}
    \hat{\textbf{I}}^i_{0|t}(p) = \frac{1}{N} \sum_{j=0}^N {\hat{\textbf{I}}^{j\rightarrow i}_{0|t}}(p),
\end{equation}
where $p$ is the position. With this, every frame overlapping with the current frame can contribute to the denoising process of the current frame. Consensus in overall structure and color appearance can be reached quickly in the early semantic reconstruction stage of the denoising process.

\paragraph{Candidates Fusion at Detail Refinement Stage.}
However, the above fusion by averaging may smooth out the high-frequency details generated during the detail refinement stage due to misalignment (Fig.~\ref{fig:misalign}). 
To generate consistent high-frequency details for corresponding regions, we propagate the generated detail with alternating sampling strategy during the detail refinement stage. We randomly anchored one stylized frame $\hat{\textbf{I}}^j_{0|t}=\textbf{I}^j_{0|t}$ at each timestep and propagate the details to overlapping regions in other frames $\hat{\textbf{I}}^i_{0|t}=\hat{\textbf{I}}^{j\rightarrow i}_{0|t}$ to overwrite conflict textures. With this pseudo-equal sharing way, we can generate consistent appearance with highly-detail fidelity. 

\section{Experimental Results}
\label{sec:result}
\subsection{Experimental Settings}

In practice, we implement our approach over stable diffusion v1-5~\cite{rombach2022high}. We use VideoFlow~\cite{shi2023videoflow} for optical flow estimation and compute the occlusion masks by forward-backward consistency check~\cite{meister2018unflow}. We choose the canny edge condition branch from ~\cite{zhang2023adding} as the structure guidance in our method. We apply our method on several videos from DAVIS~\cite{pont20172017}. The image resolution is set to 512 × 512. We employ DDPM~\cite{ho2020denoising} sampler with 20 steps. All experiments are conducted on an NVIDIA GTX3090 GPU. 
In terms of running time, a $512\times 512$ video clip with 8 frames takes about 45 seconds to generate.

\begin{figure*}[th!]
    \centering
    \settoheight{\tempdim}{\includegraphics[width=.11\linewidth]{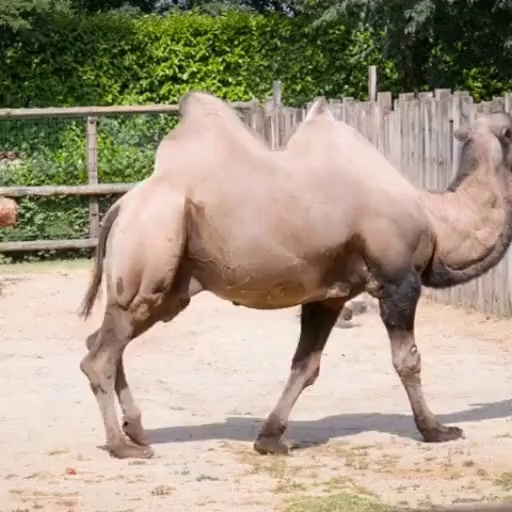}}%
    \rotatebox{90}{\makebox[\tempdim]{\small (a) Input }}\hfil
    \begin{minipage}[b]{.965\linewidth}
        \includegraphics[width=.12\linewidth]{figs/comparison/4/source/0000.jpg}
        \includegraphics[width=.12\linewidth]{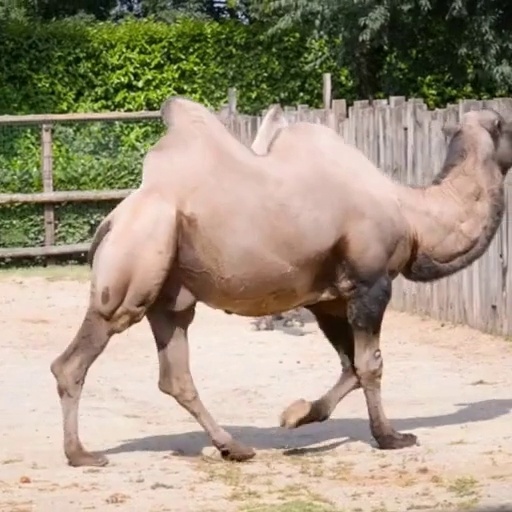}
        \includegraphics[width=.12\linewidth]{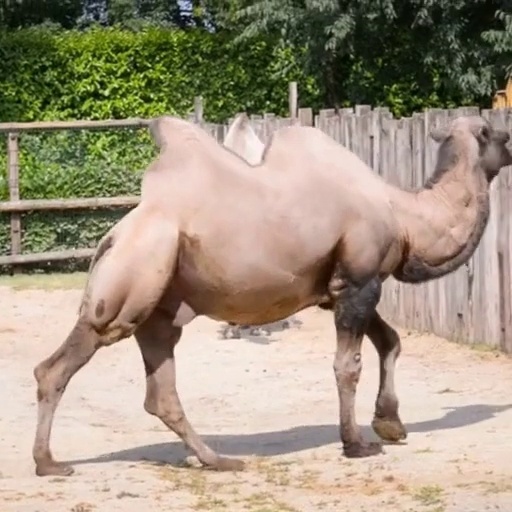}
        \includegraphics[width=.12\linewidth]{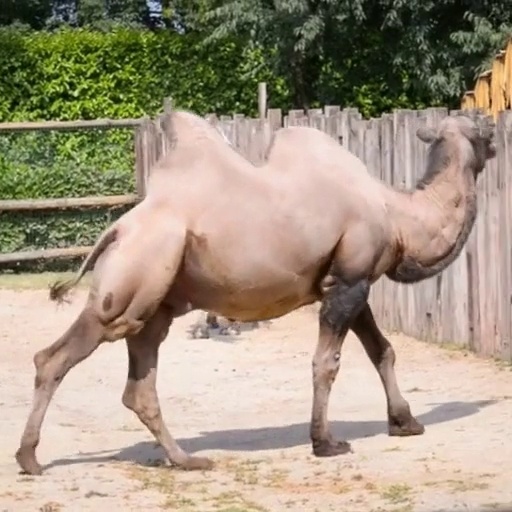}
        \includegraphics[width=.12\linewidth]{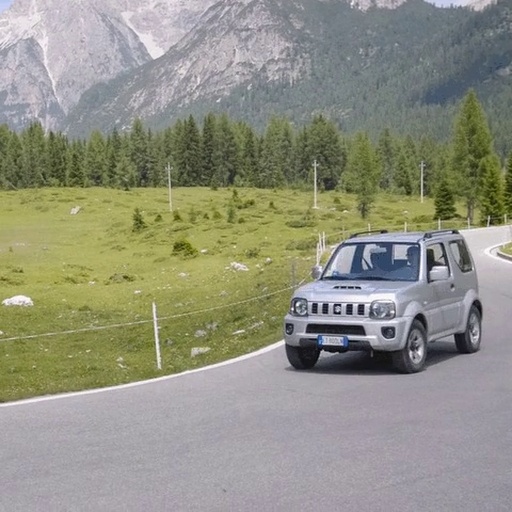}
        \includegraphics[width=.12\linewidth]{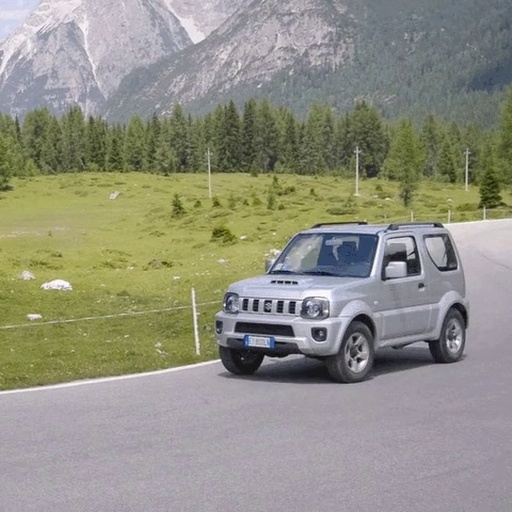}
        \includegraphics[width=.12\linewidth]{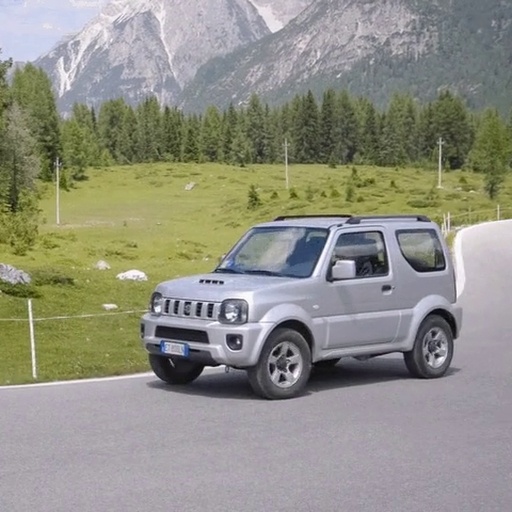}
        \includegraphics[width=.12\linewidth]{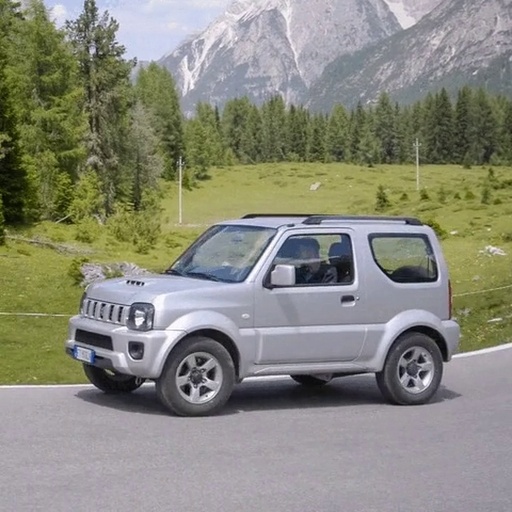}
    \end{minipage}
    \rotatebox{90}{\makebox[\tempdim]{\small (b) ControlNet }}\hfil
    \begin{minipage}[b]{.965\linewidth}
        \includegraphics[width=.12\linewidth]{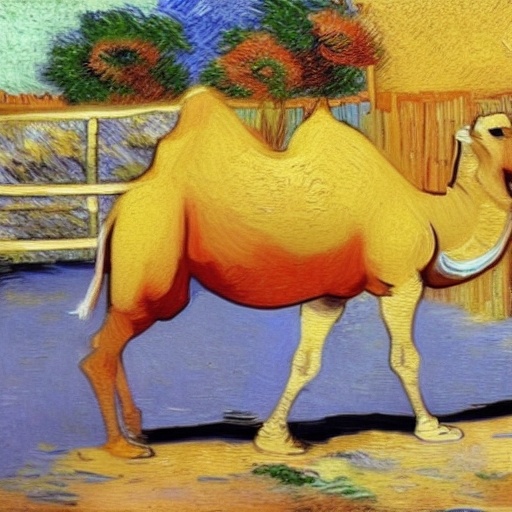}
        \includegraphics[width=.12\linewidth]{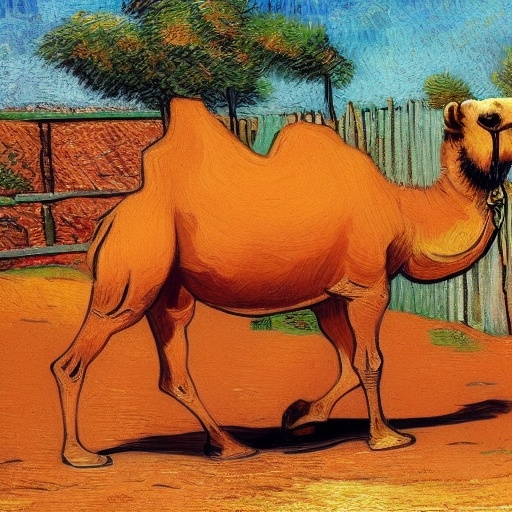}
        \includegraphics[width=.12\linewidth]{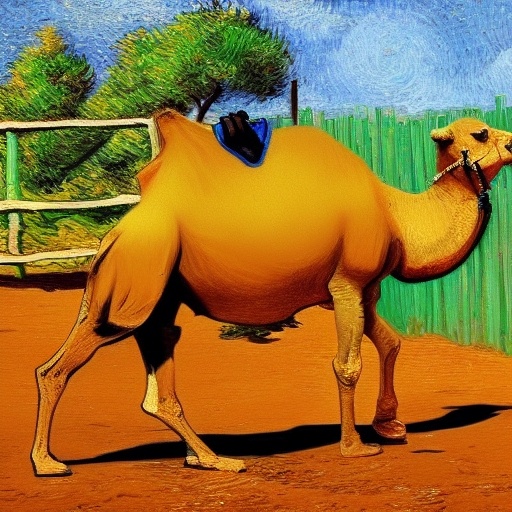}
        \includegraphics[width=.12\linewidth]{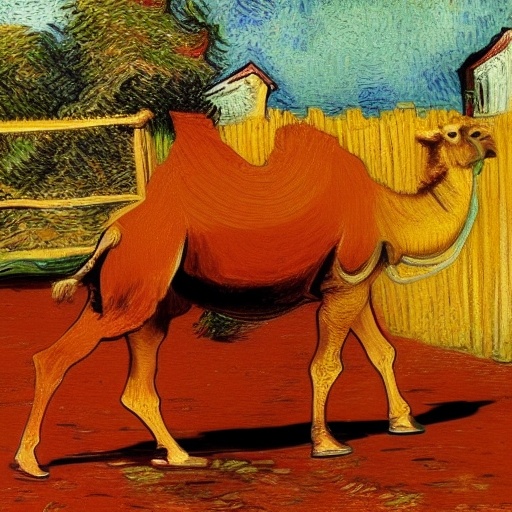}
        \includegraphics[width=.12\linewidth]{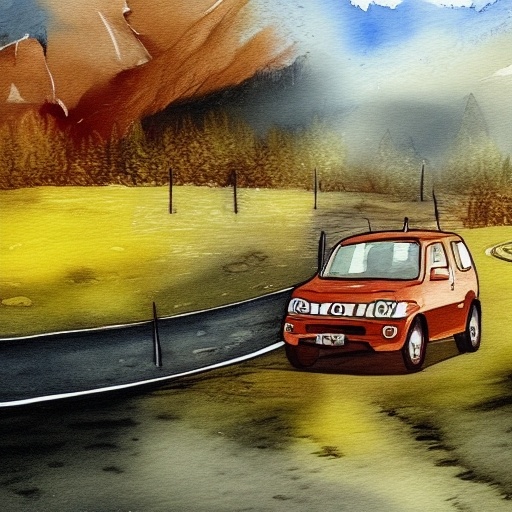}
        \includegraphics[width=.12\linewidth]{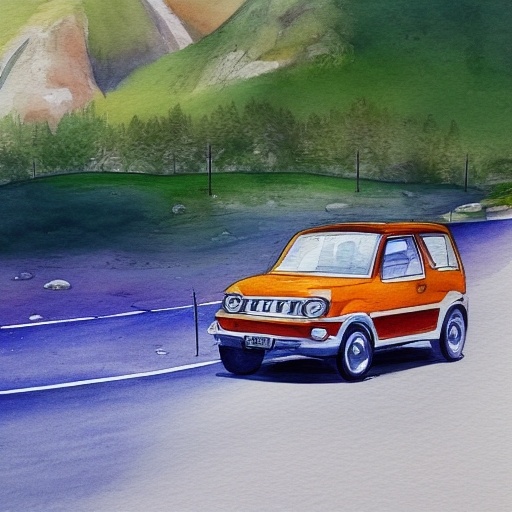}
        \includegraphics[width=.12\linewidth]{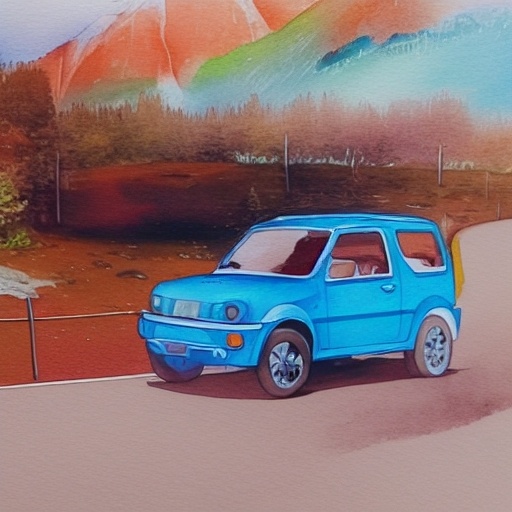}
        \includegraphics[width=.12\linewidth]{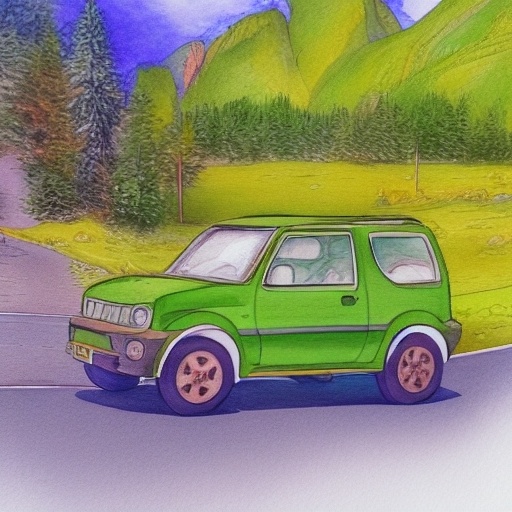}
    \end{minipage}
    \rotatebox{90}{\makebox[\tempdim]{\small (c) FateZero }}\hfil
    \begin{minipage}[b]{.965\linewidth}
        \includegraphics[width=.12\linewidth]{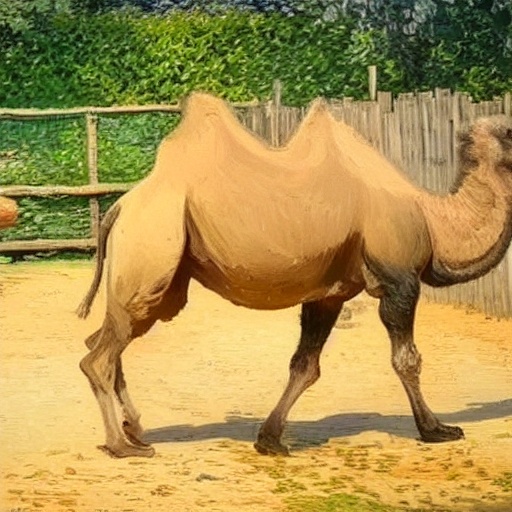}
        \includegraphics[width=.12\linewidth]{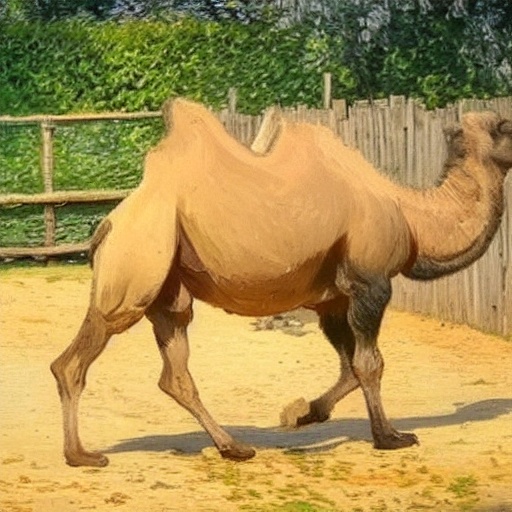}
        \includegraphics[width=.12\linewidth]{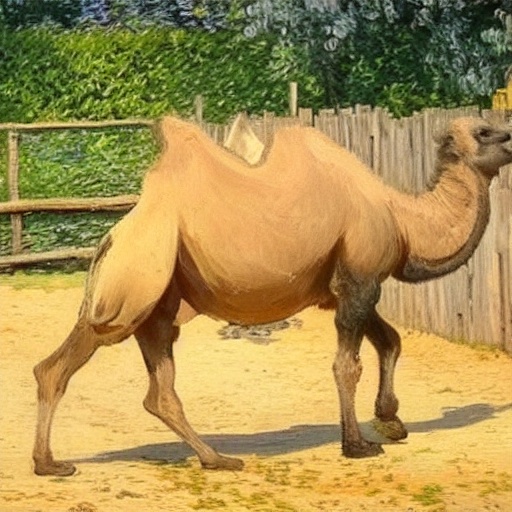}
        \includegraphics[width=.12\linewidth]{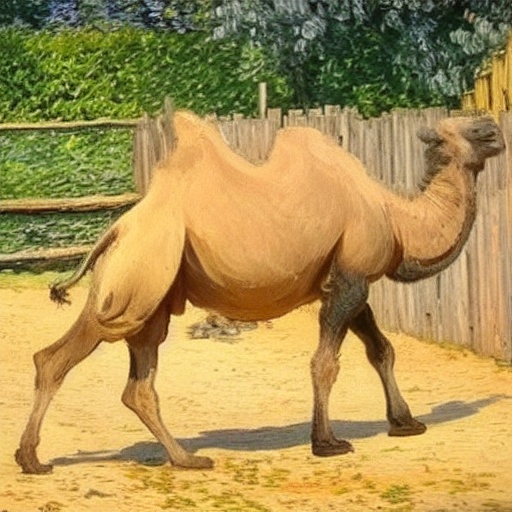}
        \includegraphics[width=.12\linewidth]{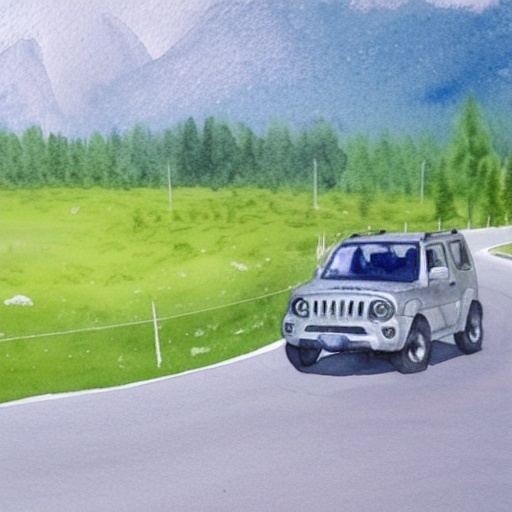}
        \includegraphics[width=.12\linewidth]{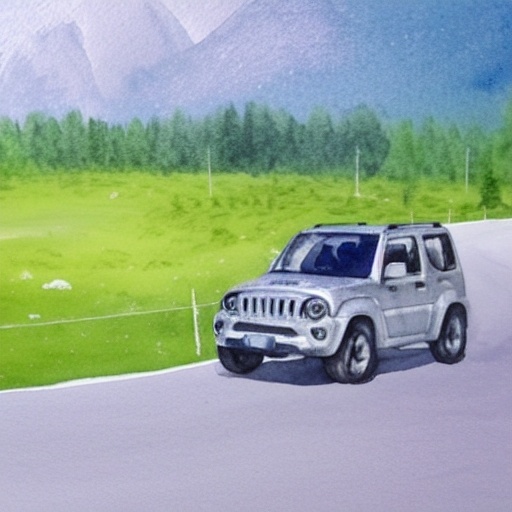}
        \includegraphics[width=.12\linewidth]{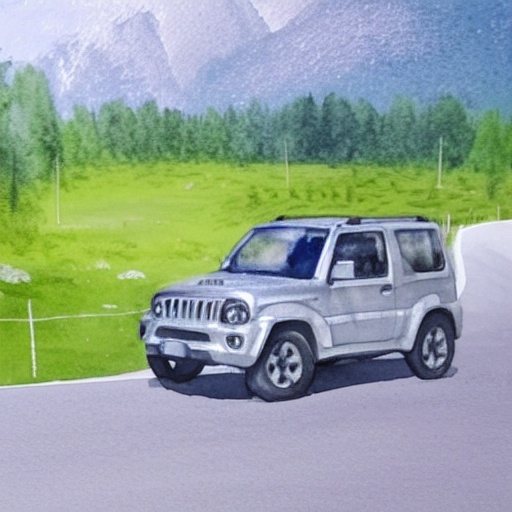}
        \includegraphics[width=.12\linewidth]{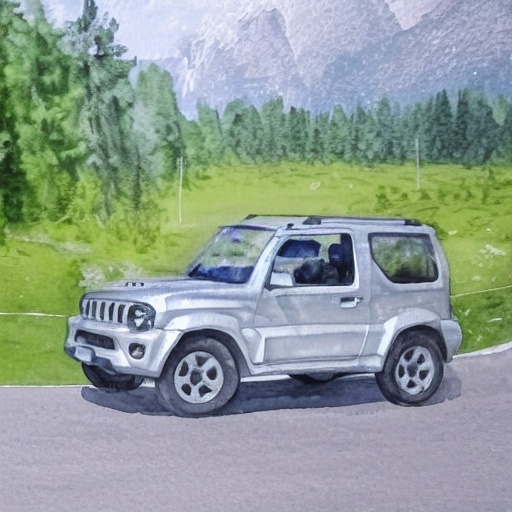}
    \end{minipage}
    \rotatebox{90}{\makebox[\tempdim]{\small (d) T2V-Zero }}\hfil
    \begin{minipage}[b]{.965\linewidth}
        \includegraphics[width=.12\linewidth]{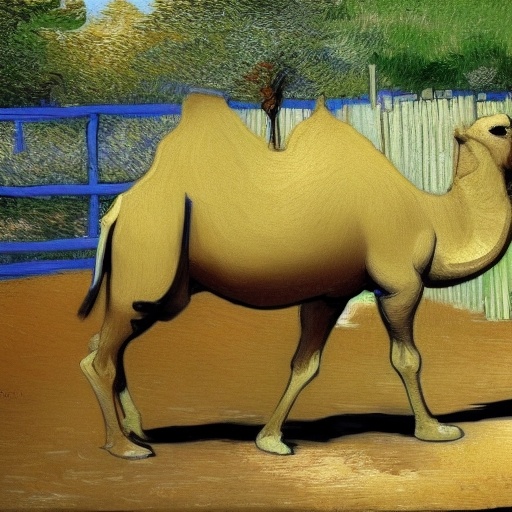}
        \includegraphics[width=.12\linewidth]{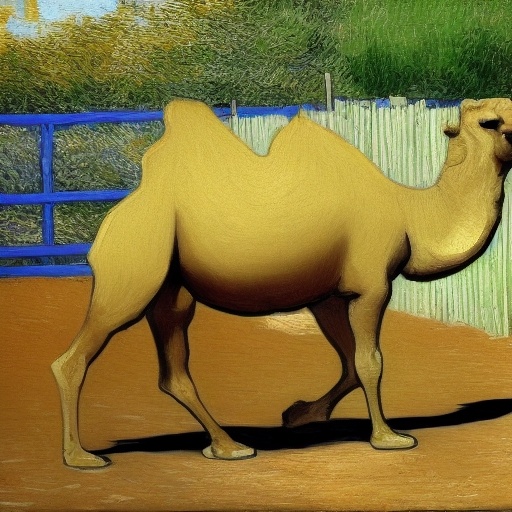}
        \includegraphics[width=.12\linewidth]{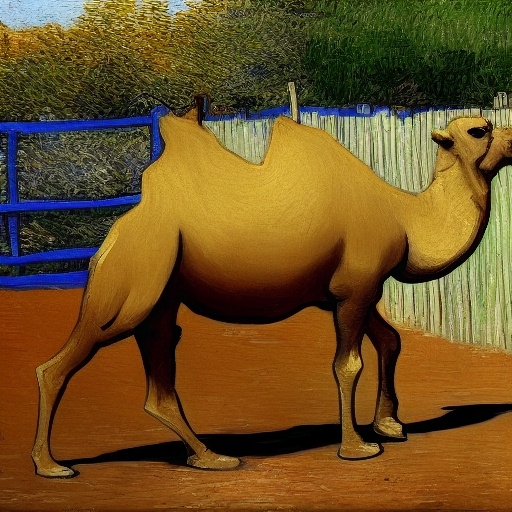}
        \includegraphics[width=.12\linewidth]{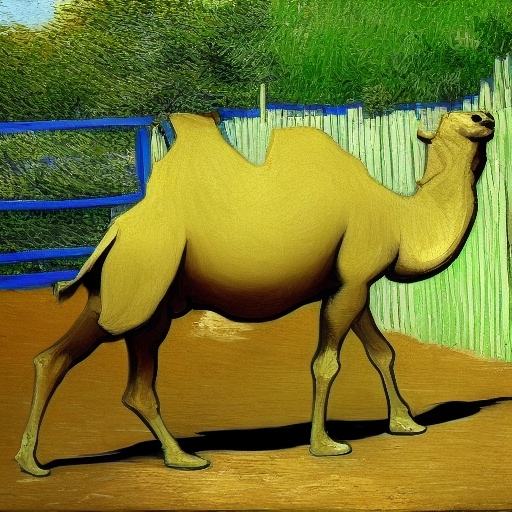}
        \includegraphics[width=.12\linewidth]{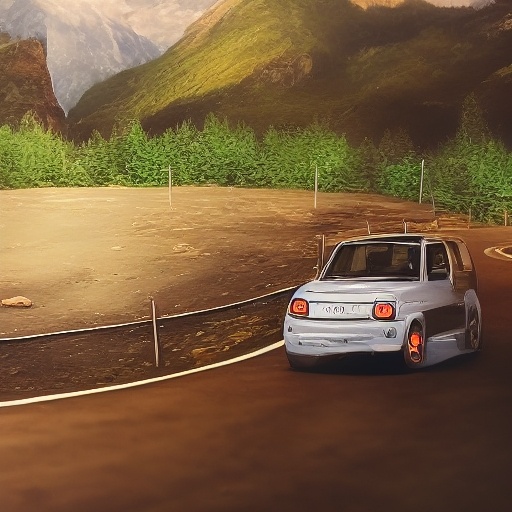}
        \includegraphics[width=.12\linewidth]{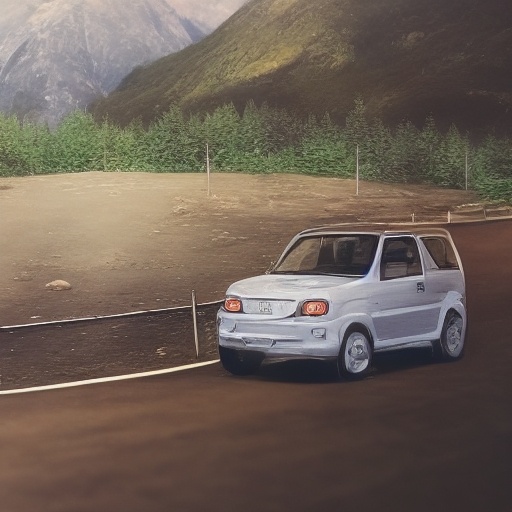}
        \includegraphics[width=.12\linewidth]{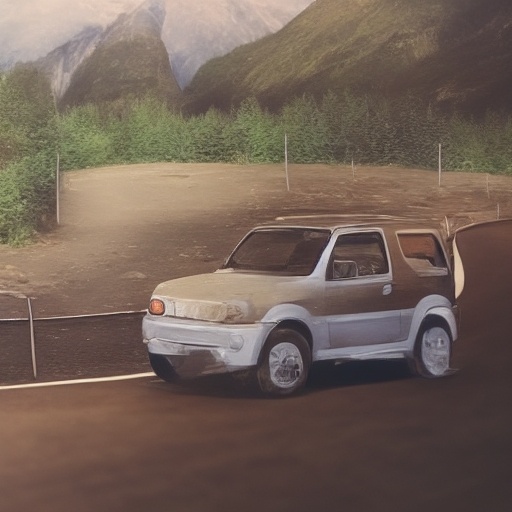}
        \includegraphics[width=.12\linewidth]{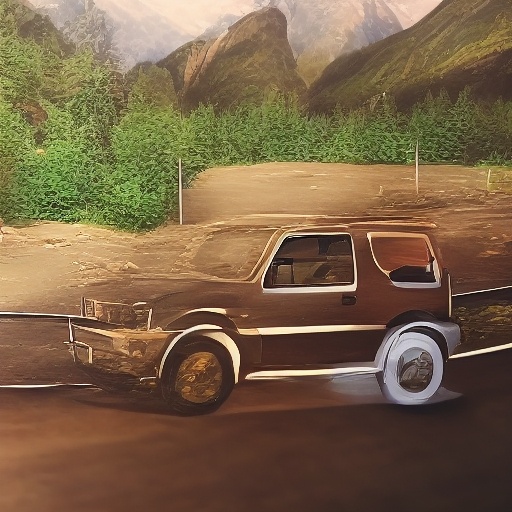}
    \end{minipage}
    \rotatebox{90}{\makebox[\tempdim]{\small (e) RAV }}\hfil
    \begin{minipage}[b]{.965\linewidth}
        \includegraphics[width=.12\linewidth]{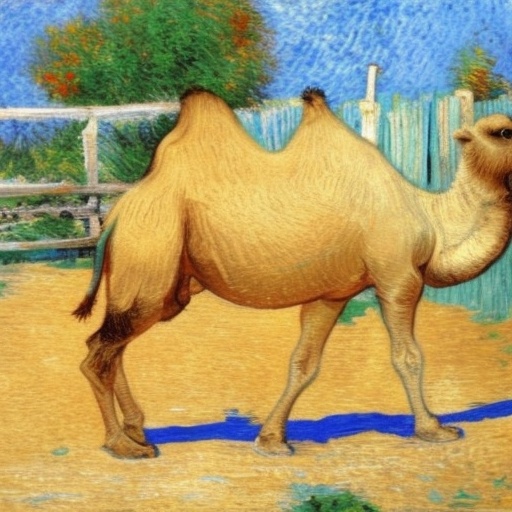}
        \includegraphics[width=.12\linewidth]{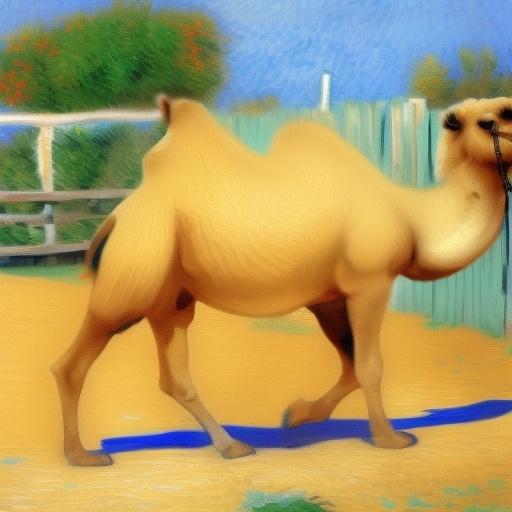}
        \includegraphics[width=.12\linewidth]{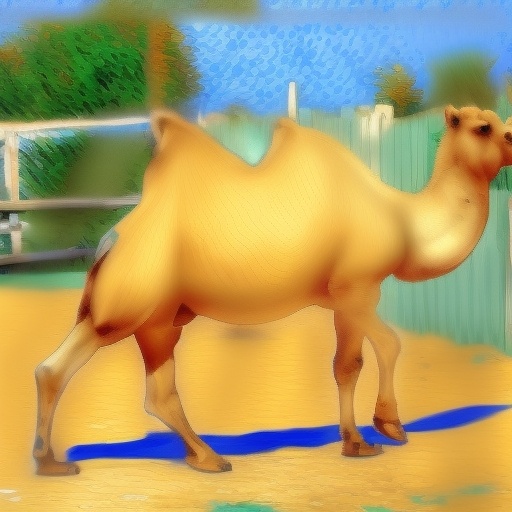}
        \includegraphics[width=.12\linewidth]{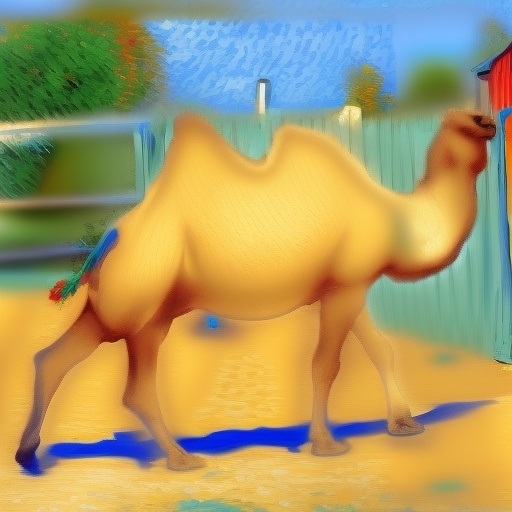}
        \includegraphics[width=.12\linewidth]{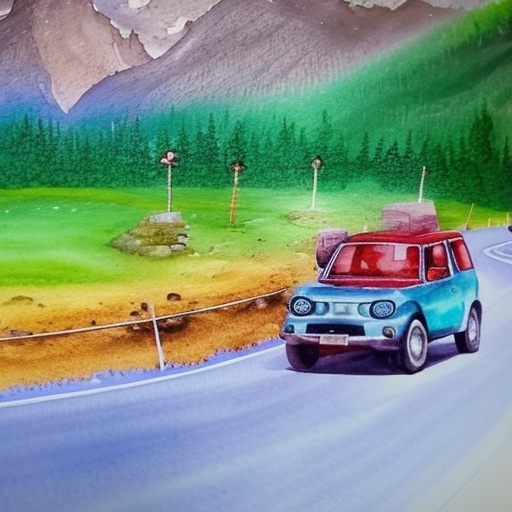}
        \includegraphics[width=.12\linewidth]{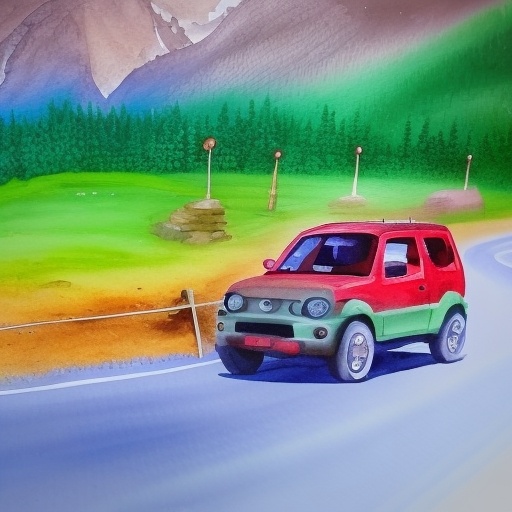}
        \includegraphics[width=.12\linewidth]{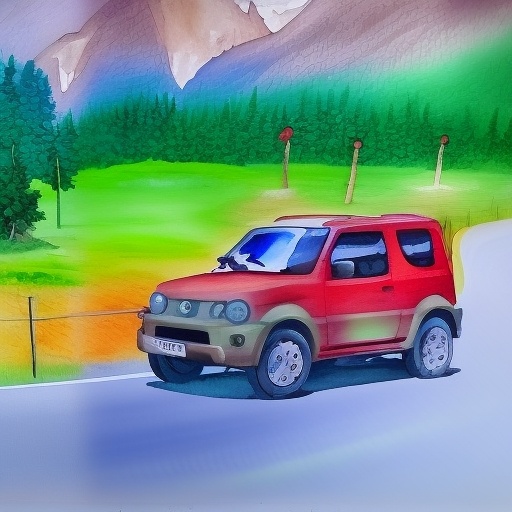}
        \includegraphics[width=.12\linewidth]{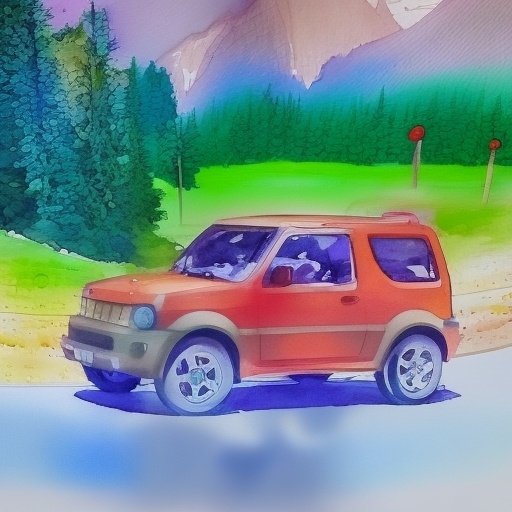}
    \end{minipage}
    \rotatebox{90}{\makebox[\tempdim]{\small (f) AD+ }}\hfil
    \begin{minipage}[b]{.965\linewidth}
        \includegraphics[width=.12\linewidth]{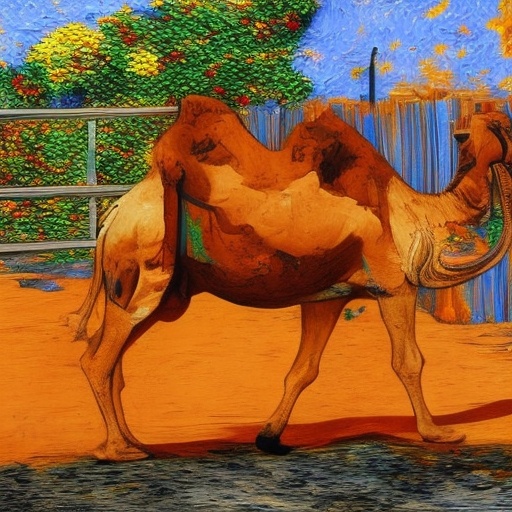}
        \includegraphics[width=.12\linewidth]{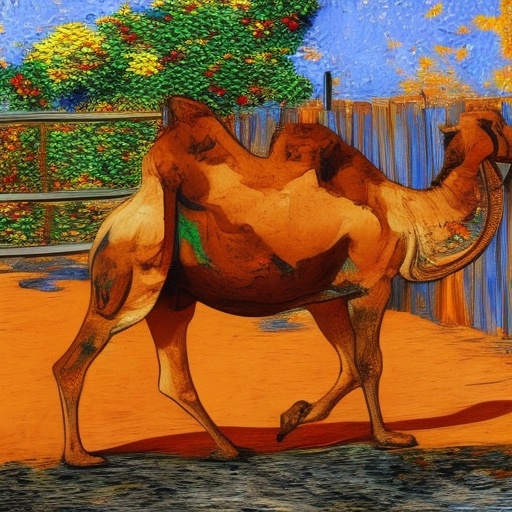}
        \includegraphics[width=.12\linewidth]{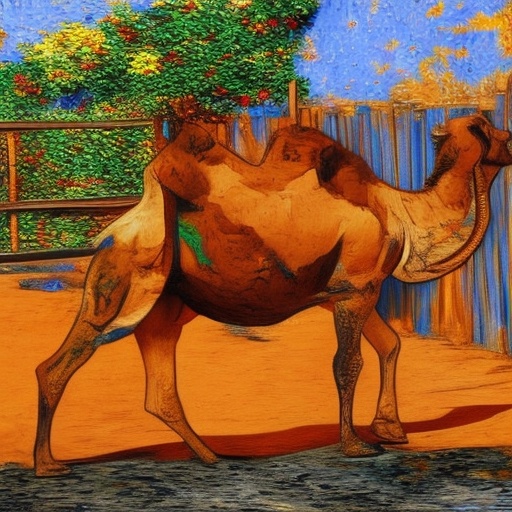}
        \includegraphics[width=.12\linewidth]{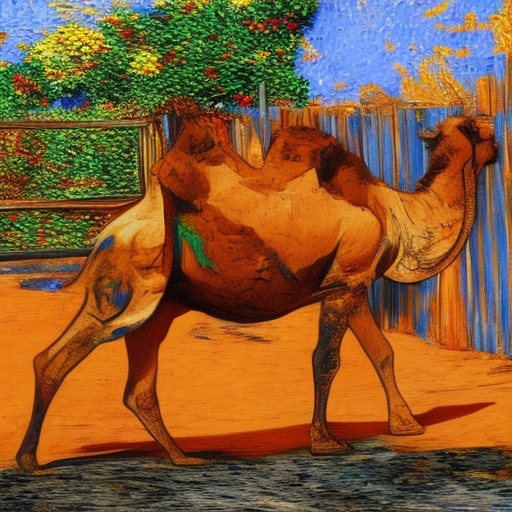}
        \includegraphics[width=.12\linewidth]{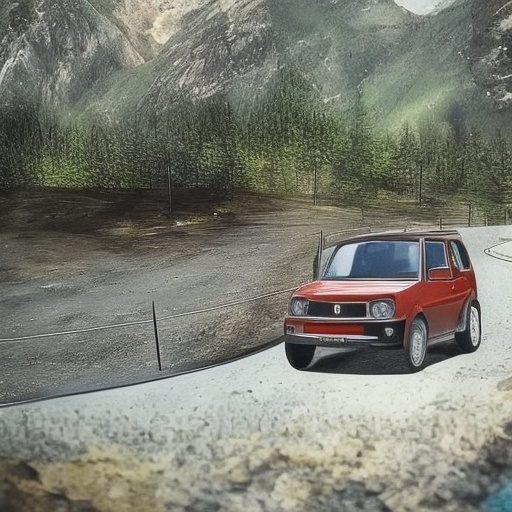}
        \includegraphics[width=.12\linewidth]{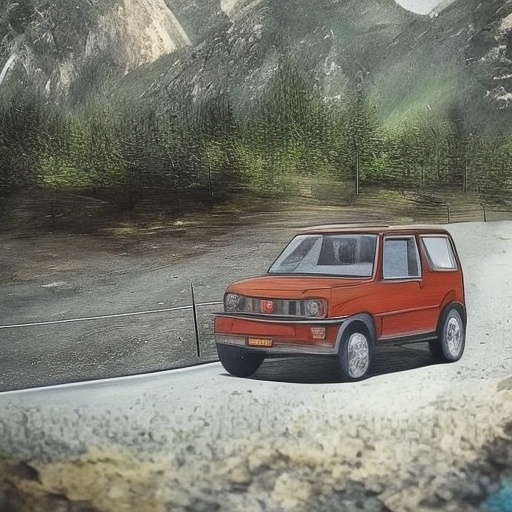}
        \includegraphics[width=.12\linewidth]{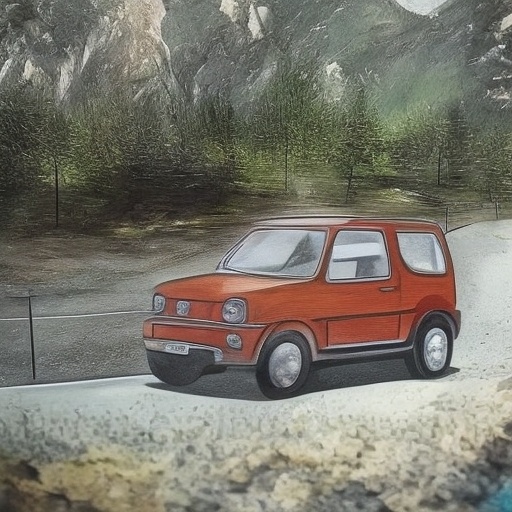}
        \includegraphics[width=.12\linewidth]{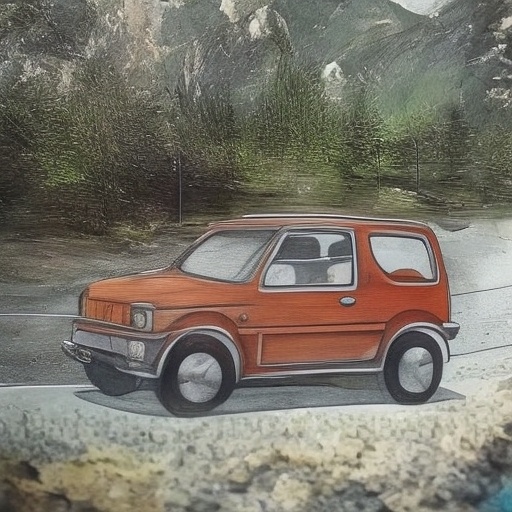}
    \end{minipage}
    \rotatebox{90}{\makebox[\tempdim]{\small (g) Ours }}\hfil
    \begin{minipage}[b]{.965\linewidth}
        \includegraphics[width=.12\linewidth]{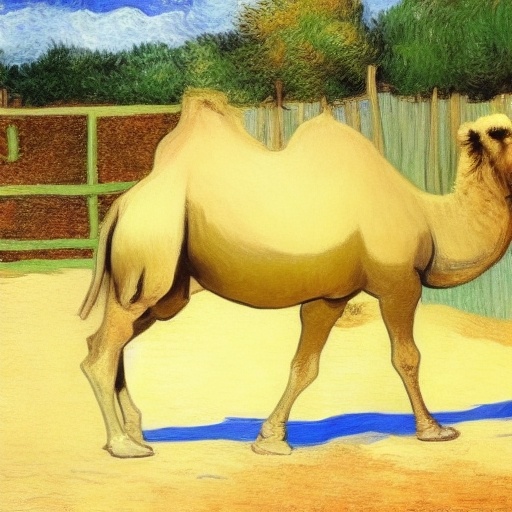}
        \includegraphics[width=.12\linewidth]{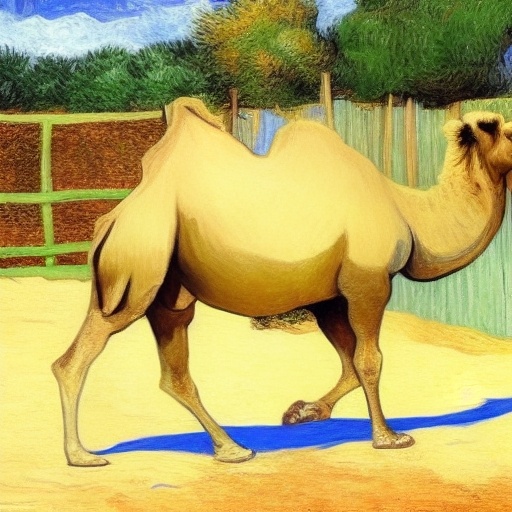}
        \includegraphics[width=.12\linewidth]{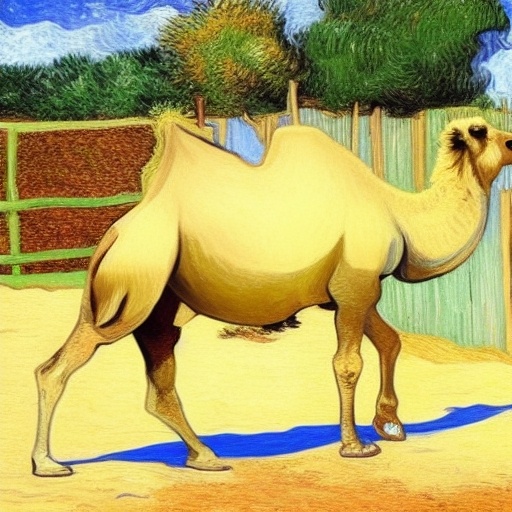}
        \includegraphics[width=.12\linewidth]{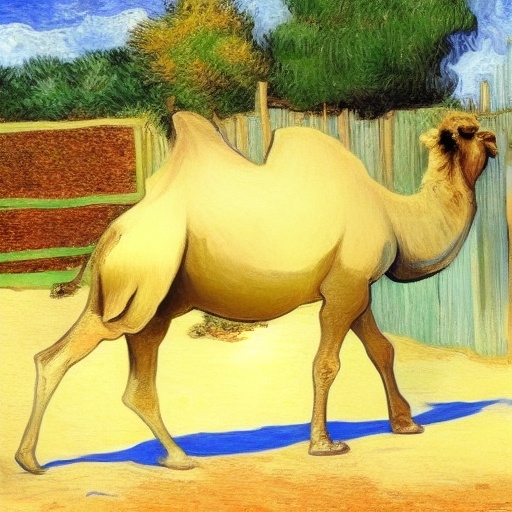}
        \includegraphics[width=.12\linewidth]{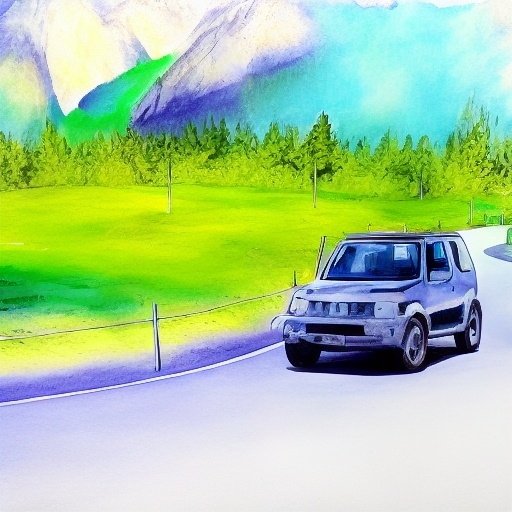}
        \includegraphics[width=.12\linewidth]{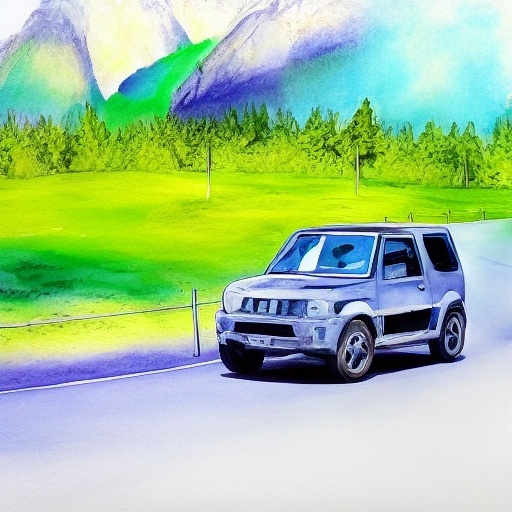}
        \includegraphics[width=.12\linewidth]{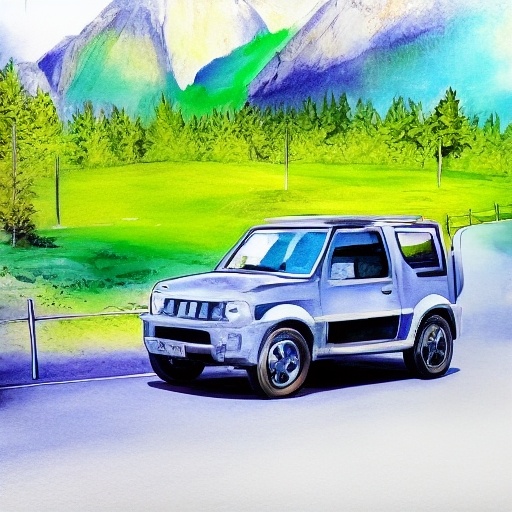}
        \includegraphics[width=.12\linewidth]{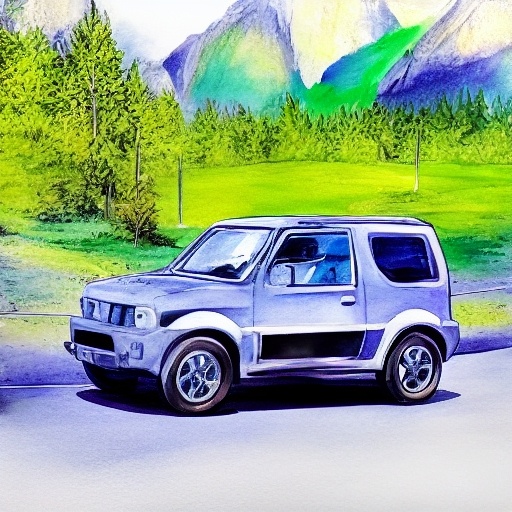}
    \end{minipage}
    \caption{Stylized results comparison. Our method can generate consistent results with more details. Text prompts: {\footnotesize\tt "A camel is walking in the dirt, Van Gogh style."} and {\footnotesize\tt "A small car driving down a road in the mountains, water coloring." } Readers are encouraged to zoom in to better compare the fine details and visual content consistency of different methods.}
    \label{fig:comparison1}
\end{figure*}

\begin{figure}[th!]
    \centering
    \settoheight{\tempdim}{\includegraphics[width=.225\linewidth]{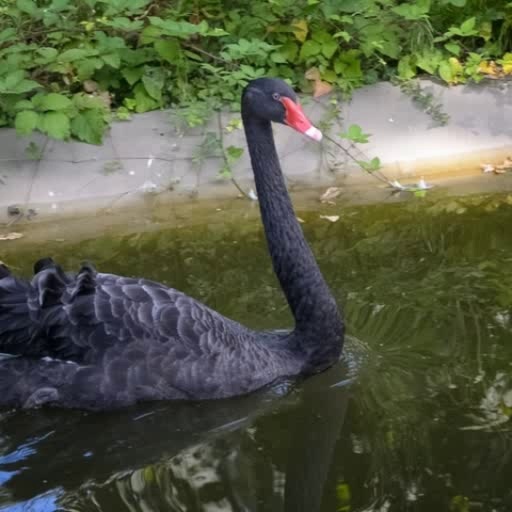}}%
    \rotatebox{90}{\makebox[\tempdim]{\small (a) Input }}\hfil
    \begin{minipage}[b]{.95\linewidth}
        \includegraphics[width=.24\linewidth]{figs/comparison/6/input/0000.jpg}
        \includegraphics[width=.24\linewidth]{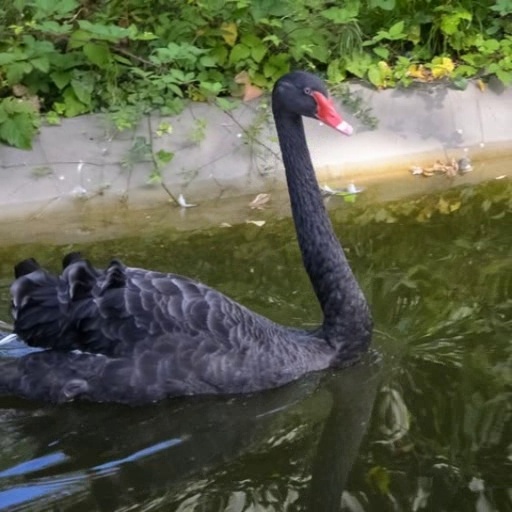}
        \includegraphics[width=.24\linewidth]{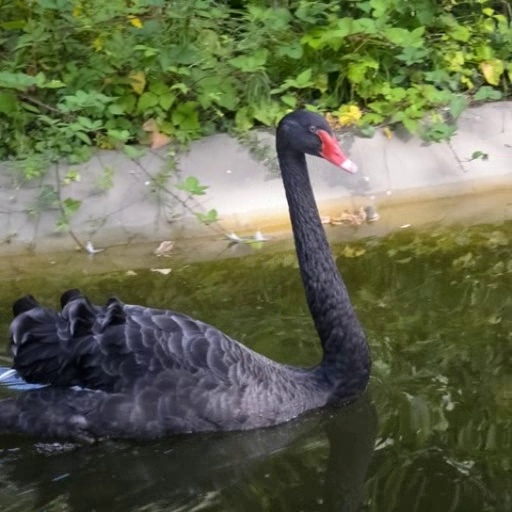}
        \includegraphics[width=.24\linewidth]{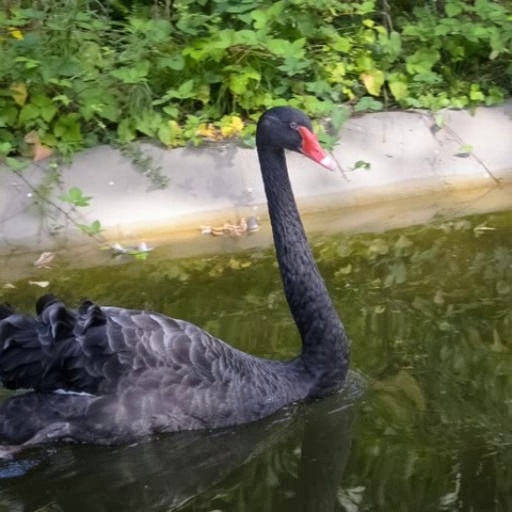}
    \end{minipage}
    \rotatebox{90}{\makebox[\tempdim]{\small (b) StableVideo }}\hfil
    \begin{minipage}[b]{.95\linewidth}
        \includegraphics[width=.24\linewidth]{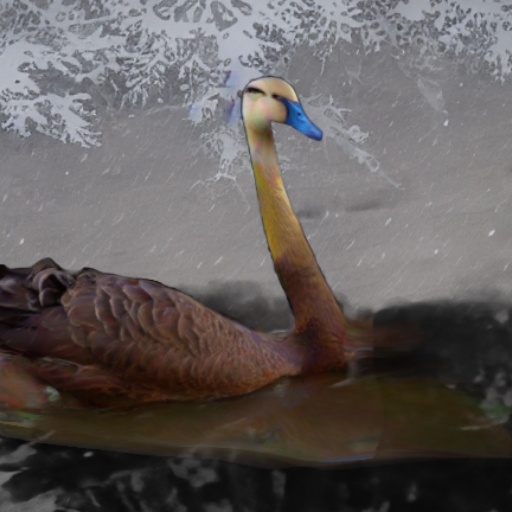}
        \includegraphics[width=.24\linewidth]{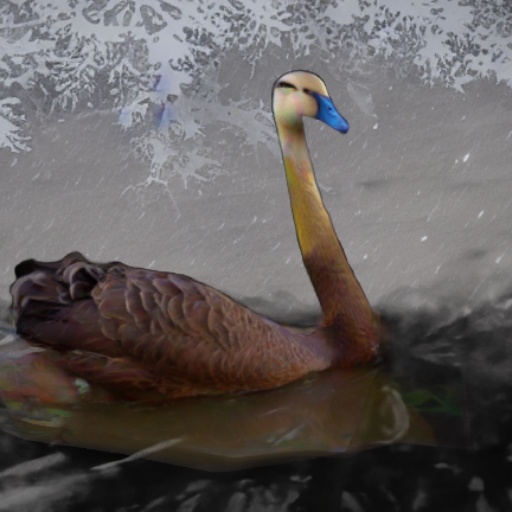}
        \includegraphics[width=.24\linewidth]{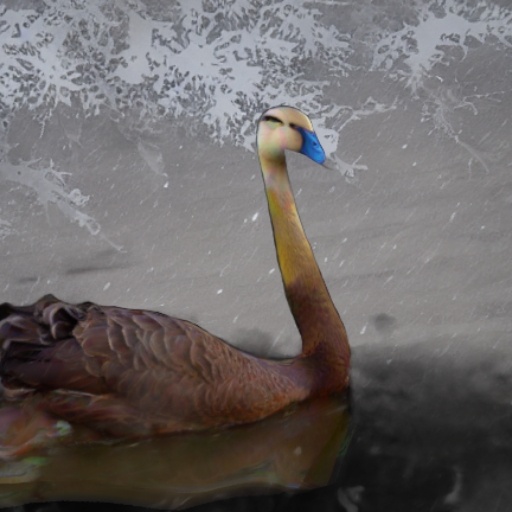}
        \includegraphics[width=.24\linewidth]{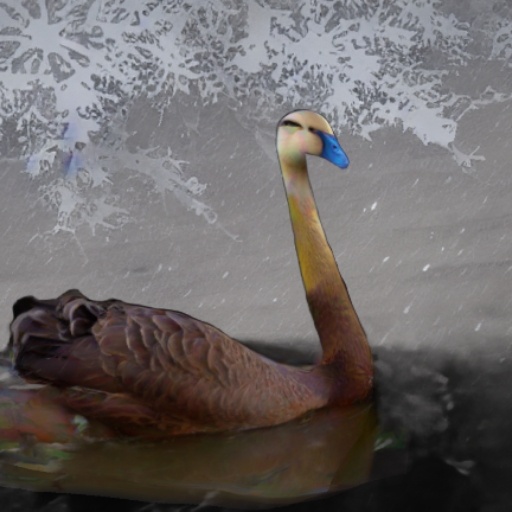}
    \end{minipage}
    \rotatebox{90}{\makebox[\tempdim]{\small (c) Ours }}\hfil
    \begin{minipage}[b]{.95\linewidth}
        \includegraphics[width=.24\linewidth]{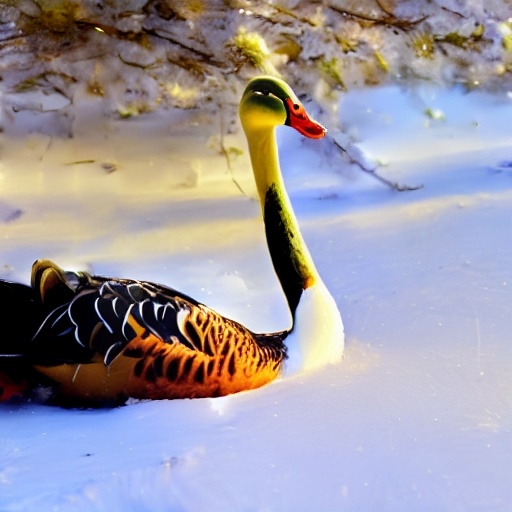}
        \includegraphics[width=.24\linewidth]{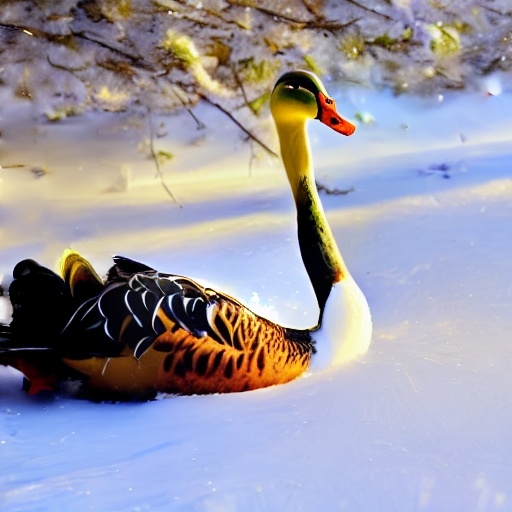}
        \includegraphics[width=.24\linewidth]{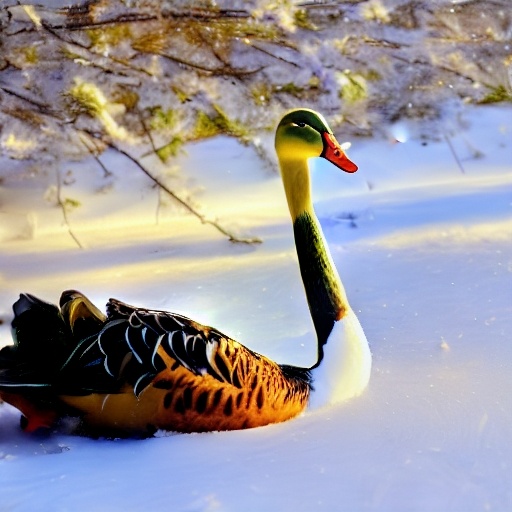}
        \includegraphics[width=.24\linewidth]{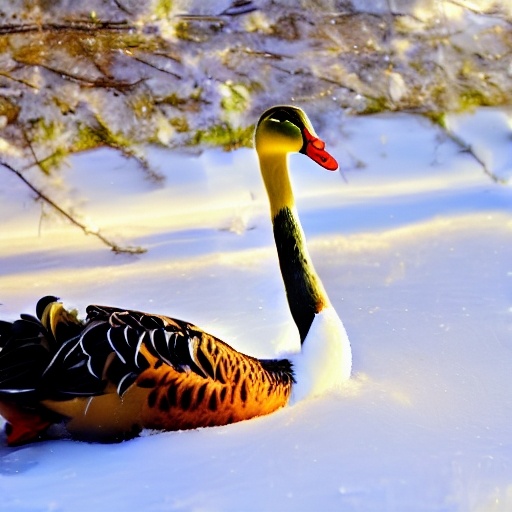}
    \end{minipage}
    \caption{Stylized results comparison to StableVideo~\cite{chai2023stablevideo}. Text prompt: {\footnotesize\tt "A duck in winter snowy scene."}}
    \label{fig:comparison2}
\end{figure}

\subsection{Comparison with State-of-the-Art Methods}

In this section, we compare our editing results with three recent zero-shot methods: FateZero~\cite{qi2023fatezero}, Text2Video-Zero (T2V-Zero)~\cite{khachatryan2023text2video}, and Rerender-A-Video (RAV)~\cite{yang2023rerender}, and two methods with extra training: AnimateDiff (AD)~\cite{guo2023animatediff} and StableVideo~\cite{chai2023stablevideo}. 
Besides, we also select ControlNet~\cite{zhang2023adding} as a competitor to evaluate the geometric constraint. As the official code of AnimateDiff~\cite{guo2023animatediff} does not support ControlNet~\cite{zhang2023adding}, it fails to generate video with similar geometry as the original video. Thus, we re-implement it to support ControlNet~\cite{zhang2023adding} for comparison, named AnimateDiff+. 

Figures~\ref{fig:comparison1} and~\ref{fig:comparison2} present the visual results.
FateZero~\cite{qi2023fatezero} will fail to edit the input video when it fail to extract correct cross-attention map for the user text prompt, leading to stylized frames similar to the input video. 
While each frame generated by Text2Video-Zero~\cite{khachatryan2023text2video} is of high quality and generate consistent global style, they may suffer from color jittering and lack of consistency in medium- and fine-scale details/texture. Because Rerender-A-Video~\cite{yang2023rerender} follows a continuous generation, stylized frames may suffer from over-blurring in later frames (readers are encouraged to blow up the figure for better inspection). AnimateDiff+ can produce frames with rich textures, but it does not follow the motion of the original movie. For example in the Fig.~\ref{fig:comparison1}(f) camel example, the panned background in the original video becomes static in their stylized output. This negligence of motion is also reflected in our quantitative evaluation of temporal consistency in Table~\ref{tab:evaluation} (metrics Mont-MSE).
Although StableVideo~\cite{chai2023stablevideo} can produce temporally consistent video, it can produce noticeable seams along background and foreground objects (Fig.~\ref{fig:comparison2}).
In contrast, our proposed method shows clear superiority on generating frames with temporal consistency and clear texture details. 

\begin{table}[t!]
    \centering
    \caption{Quantitative comparison. The best score in \textbf{bold} and the first runner-up  with \underline{underline}.}
    {\small
    \begin{tabular}{c|ccc}
        Methods & Fram-Acc $\uparrow$ & Feat-Con $\uparrow$ & Mont-MSE $\downarrow$\\ \hline
        StableDiffusion & 0.9104 & 0.8545 & 167.5751 \\  
        Controlnet & 0.7478 & 0.8828 & 93.1104 \\  \hline
        FateZero &  0.2133 & \textbf{0.9814} & \textbf{12.0448}  \\  
        T2V-Zero & 0.7502 & 0.9443 & 50.2440 \\  
        Rerender-A-Video & 0.5319 & 0.9556 & 43.6998 \\  
        AnimateDiff+ & \textbf{0.7940} & 0.9707 & 23.3980 \\  
        Ours & \underline{0.7891} & \underline{0.9785} & \underline{20.0540} \\  
    \end{tabular}
    }
    \label{tab:evaluation}
\end{table}

For quantitative evaluation, we follow other methods~\cite{qi2023fatezero,ceylan2023pix2video,yang2023rerender} to compute CLIP-based frame-wise editing accuracy (Fram-Acc), and CLIP-based frame-wise cosine similarity between consecutive frames (Feat-Con). 
Fram-Acc evaluates whether the generated frames align with the target text prompt, while the Feat-Con evaluates whether consecutive frames shares similar image features.
Additionally, we employ the motion consistency of dense optical flow (Mont-MSE) of the edited video frames from StableVideo~\cite{chai2023stablevideo}. The Farneback algorithm~\cite{farneback2003two} in OpenCV~\cite{opencv_library} is employed to calculate the average L2 distance of dense optical flow between the edited and original videos. 
We manually collect 50 video clips, each with 8 frames, and generate stylized videos with 11 artistic styles, e.g. water coloring style, oil painting style, Chinese ink painting style, Pixar style, etc.  We additionally compare with the pretrained T2I diffusion model~\cite{rombach2022high} for baseline. As StableVideo requires extra training for compressed representation of a video, we did not quantitaively compare it due to the limited resource.

Table~\ref{tab:evaluation} lists the evaluation scores. As results of FateZero~\cite{qi2023fatezero} closely resemble the input video and may ignore the user text prompt, the method therefore obtains the lowest Fram-Acc score.
On the other hand, although AnimateDiff+ highly respects the user text prompt and obtains the highest Fram-Acc score, it receives a lower Feat-Con and Mont-MSE scores, i.e. weaker temporal consistency, as it sometimes ignores the motion of the original video, as demonstrated by the relatively static background in their camel and car results of Fig.~\ref{fig:comparison1}(f). 
In sharp contrast, our method highly respects the user prompt (first runner-up Fram-Acc), and faithfully follows the motion in the input video and never comes up with a static background (first runner-up in both temporal consistency scores Feat-Con and Mont-MSE). Note that even FateZero obtains highest Feat-Con and Mont-MSE, it is too similar to the input video to be useful.
In other words, our method strikes a nice balance in both the semantic conformity to the user prompt and the motion of the input video, while producing highly detailed texture content.

\subsection{Ablation Study}
\begin{figure}
    \centering
    \settoheight{\tempdim}{\includegraphics[width=.225\linewidth]{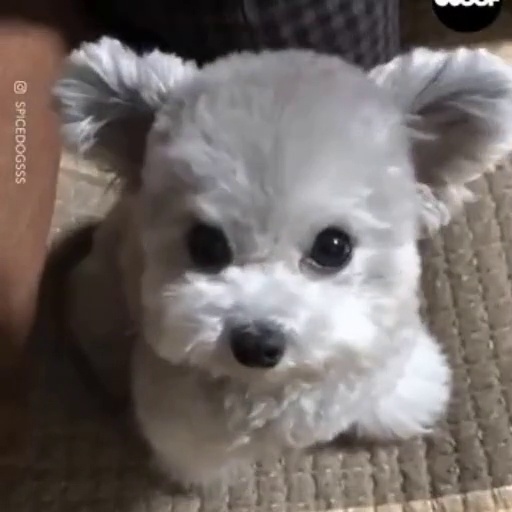}}%
    \rotatebox{90}{\makebox[\tempdim]{\small (a) Input }}\hfil
    \begin{minipage}[b]{.95\linewidth}
        \includegraphics[width=.24\linewidth]{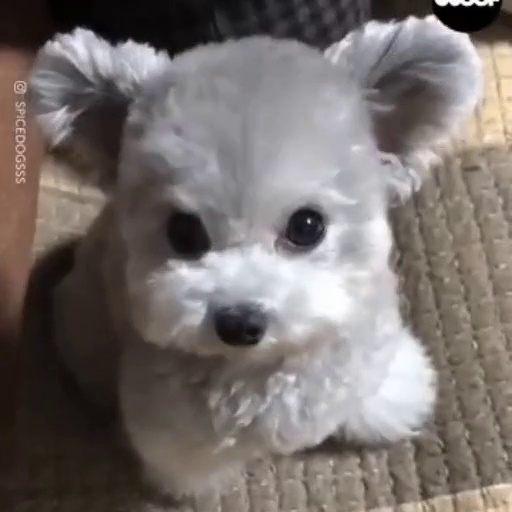}
        \includegraphics[width=.24\linewidth]{figs/ablation/mffm/input/0003.jpg}
        \includegraphics[width=.24\linewidth]{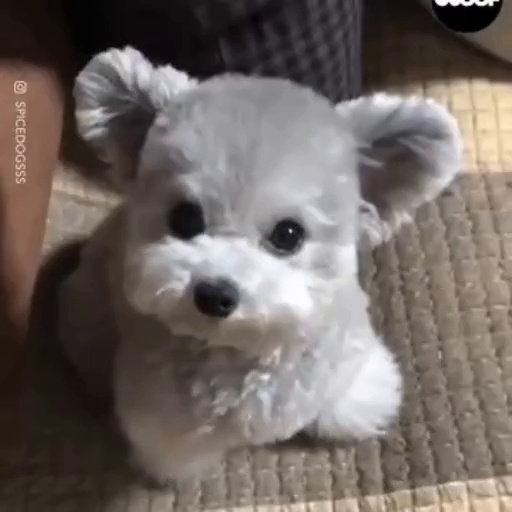}
        \includegraphics[width=.24\linewidth]{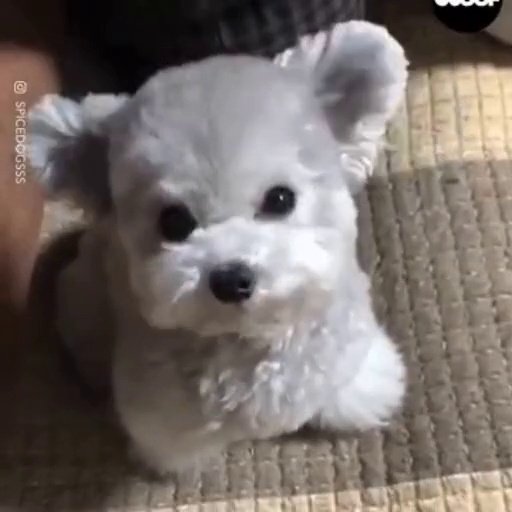}
    \end{minipage}
    \rotatebox{90}{\makebox[\tempdim]{\small (b) w/o MFFM}}\hfil
    \begin{minipage}[b]{.95\linewidth}
        \includegraphics[width=.24\linewidth]{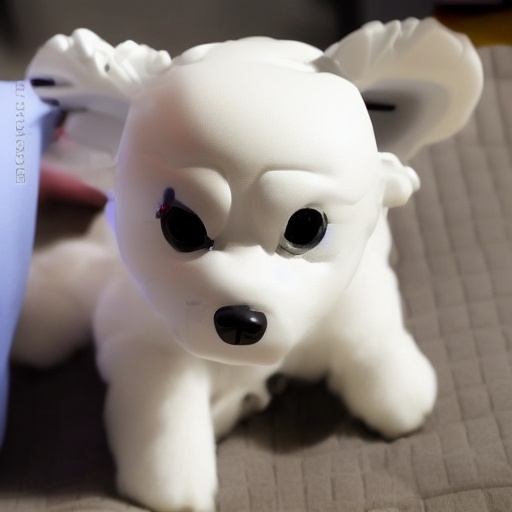}
        \includegraphics[width=.24\linewidth]{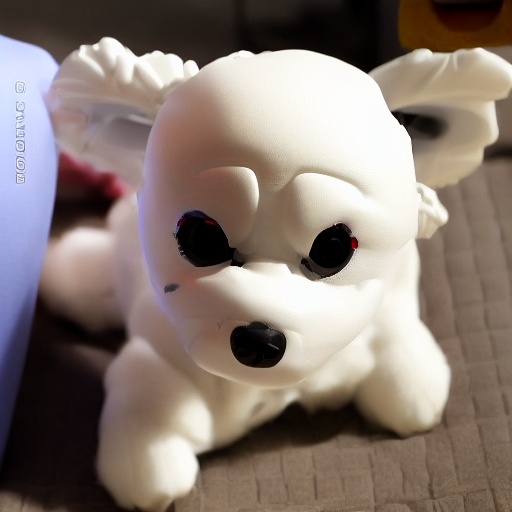}
        \includegraphics[width=.24\linewidth]{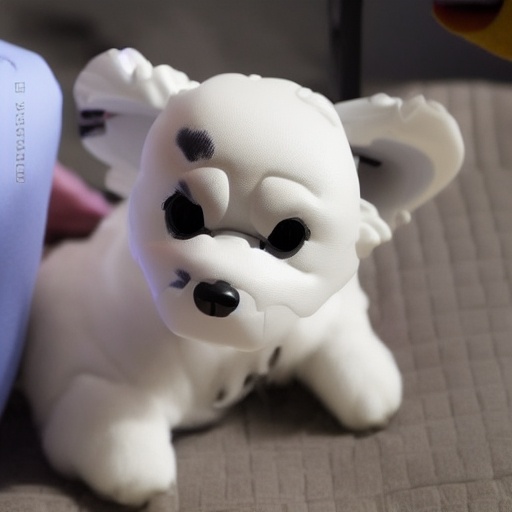}
        \includegraphics[width=.24\linewidth]{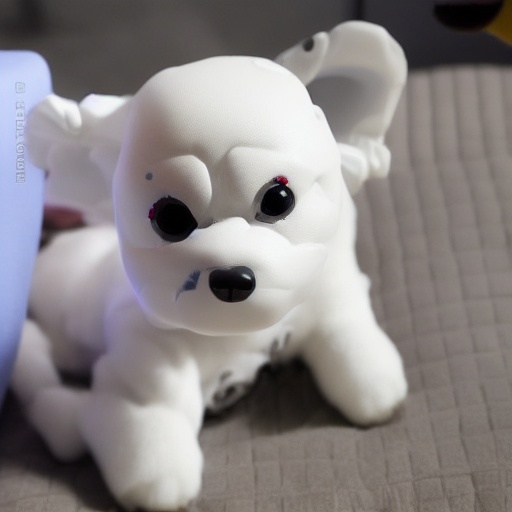}
    \end{minipage}
    \rotatebox{90}{\makebox[\tempdim]{\small (c) w MFFM}}\hfil
    \begin{minipage}[b]{.95\linewidth}
        \includegraphics[width=.24\linewidth]{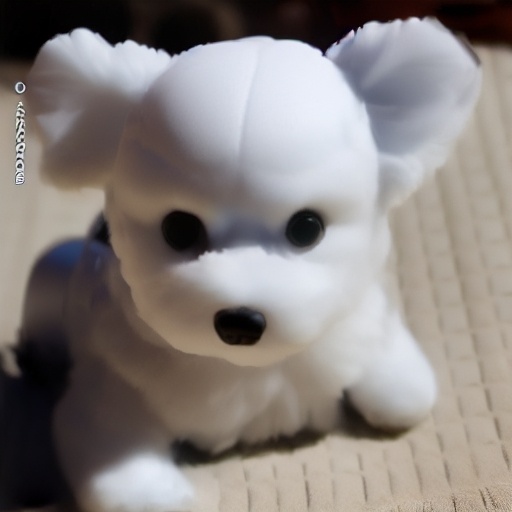}
        \includegraphics[width=.24\linewidth]{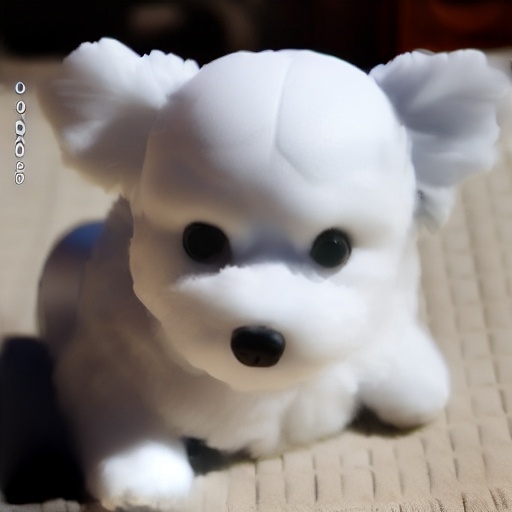}
        \includegraphics[width=.24\linewidth]{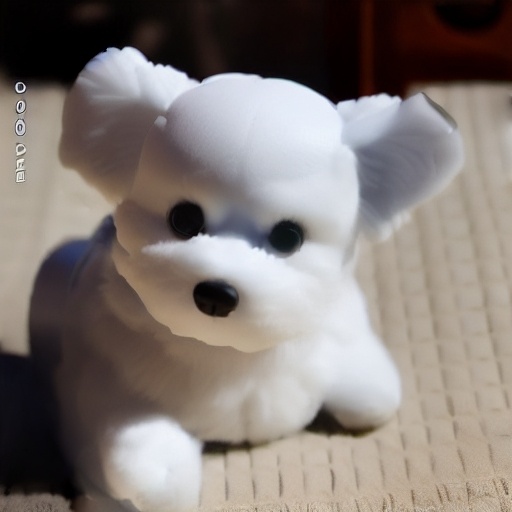}
        \includegraphics[width=.24\linewidth]{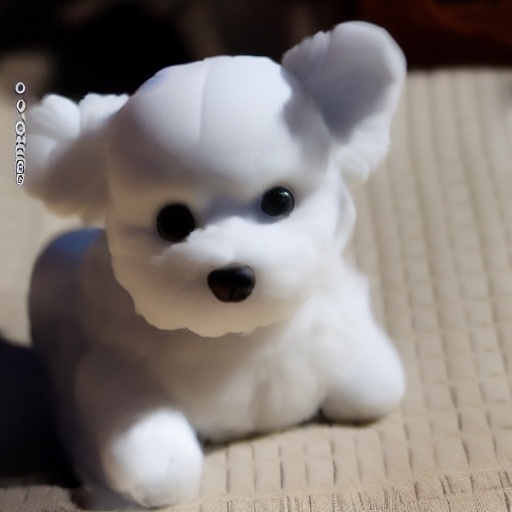}
    \end{minipage}
    \caption{Ablation study of Multi-Frame Fusion Module (MFFM). Without MFFM, the appearances of frames are very inconsistent. This evidences the importance
    of information sharing via our MFFM. Text prompt: {\footnotesize\tt "A robotic dog."}}
    \label{fig:mffm}
\end{figure}

\paragraph{Multi-Frame Fusion Module}
As the core of our research, we evaluate the impact of the Multi-Frame Fusion Module. Its objective is to allow information sharing among frames, and hence, ensure the  visual consistency among  frames in all scales. Fig.~\ref{fig:mffm} shows an example, where color and structure inconsistency exists without our proposed Multi-Frame Fusion Module.

\begin{figure}
    \centering
    \begin{subfigure}{0.48\linewidth}
        \includegraphics[width=.655\linewidth]{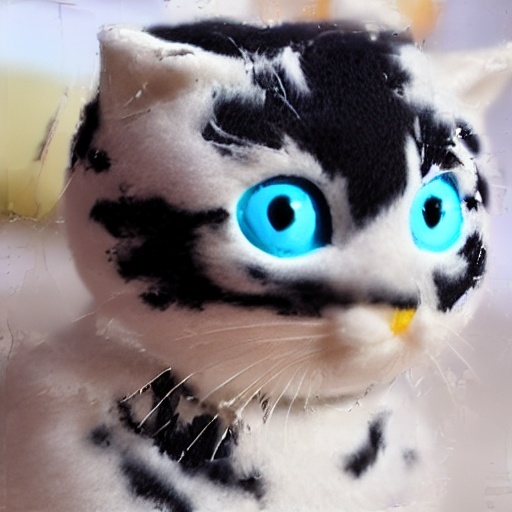}
        \begin{minipage}[b]{.325\linewidth}
            \includegraphics[width=\linewidth]{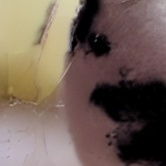}\\
            \includegraphics[width=\linewidth]{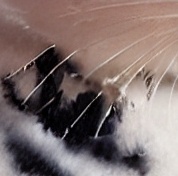}
        \end{minipage}
        \caption{w/o PIE}
    \end{subfigure}
    \begin{subfigure}{0.48\linewidth}
        \includegraphics[width=.655\linewidth]{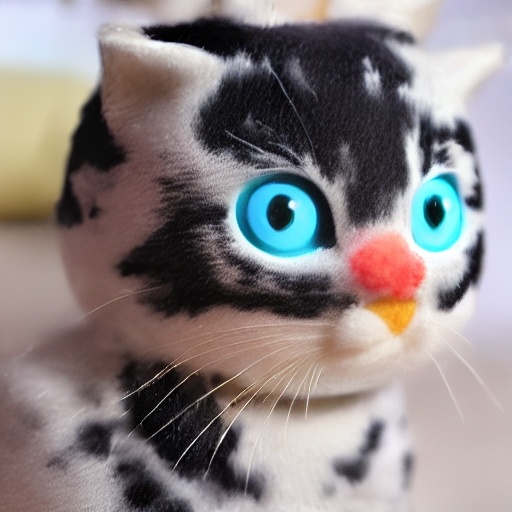}
        \begin{minipage}[b]{.325\linewidth}
            \includegraphics[width=\linewidth]{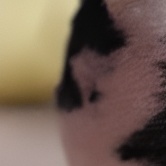}\\
            \includegraphics[width=\linewidth]{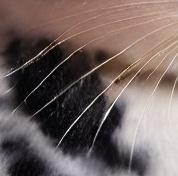}
        \end{minipage}
        \caption{w PIE}
    \end{subfigure}
    \vspace{-.05in}
    \caption{Ablation study of Poisson Image Editing (PIE). Poisson solving effectively avoids the obvious seams/fragmentation at the overlapping regions.  
    Text prompt: {\footnotesize\tt "A detailed woolen toy cat."}}
    \vspace{-.05in}
    \label{fig:poisson}
\end{figure}

\paragraph{Poisson Image Editing}
Fig.~\ref{fig:poisson} illustrates the effectiveness of Poisson solver in blending candidates to achieve information sharing across frames. For evaluation, we generate candidate regions by directly merging overlapping regions with disoccluded regions. We can see that there are noticeable seams in the final results. This is because the generated appearance of the cat between two frames may not match, leading to abrupt intensity changes along the merged boundaries. 

\begin{figure}
    \centering
    \begin{subfigure}{0.48\linewidth}
        \includegraphics[width=.655\linewidth]{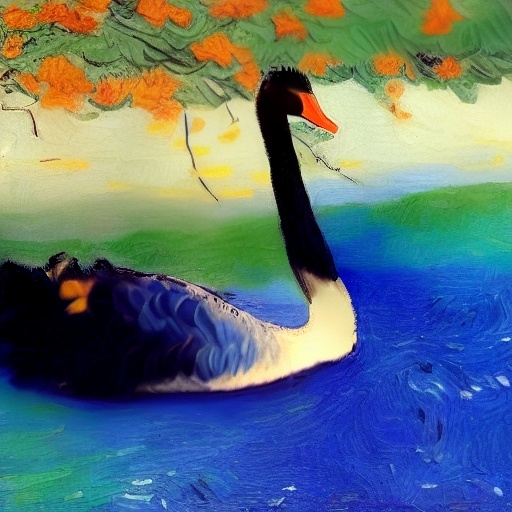}
        \begin{minipage}[b]{.325\linewidth}
            \includegraphics[width=\linewidth]{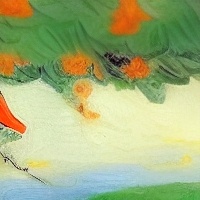}\\
            \includegraphics[width=\linewidth]{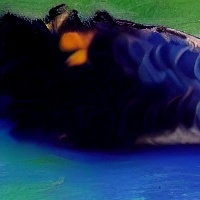}
        \end{minipage}
        \caption{w/o ADP}
    \end{subfigure}
    \begin{subfigure}{0.48\linewidth}
        \includegraphics[width=.655\linewidth]{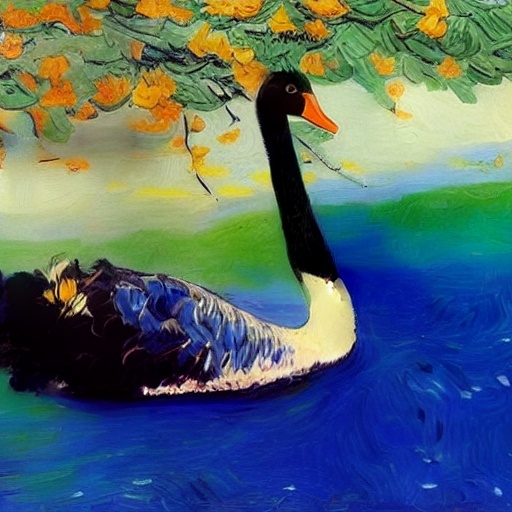}
        \begin{minipage}[b]{.325\linewidth}
            \includegraphics[width=\linewidth]{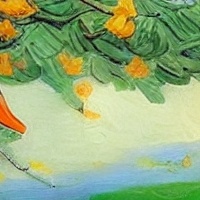}\\
            \includegraphics[width=\linewidth]{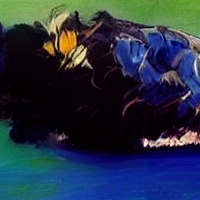}
        \end{minipage}
        \caption{w ADP}
    \end{subfigure}
    \vspace{-.05in}
    \caption{Ablation study of alternating detail propagation (ADP). Without  ADP, fine details are smoothened out at overlapping regions. Text prompt: {\footnotesize\tt "A black swan is swimming on the water, Van Gogh style."}}
    \vspace{-.05in}
    \label{fig:partial}
\end{figure}

\paragraph{Alternating Detail Propagation}
In addition, we also conducted experiments on the alternating detail propagation as shown in Fig.~\ref{fig:partial}. Merging all candidates can guarantee consistency but it may smooth out the fine details when conflict textures appear among frames during denoising steps. We can see that the feathers of the swan and flower in the background are blurred. In contrast, our pesudo-sharing strategy can help generate consistent appearance across frames while preserving the high-frequency details. 

\subsection{Limitations}
\begin{figure}
    \centering
    \settoheight{\tempdim}{\includegraphics[width=.225\linewidth]{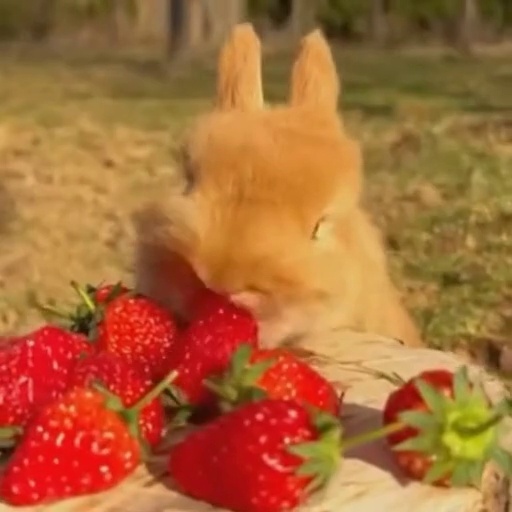}}%
    \rotatebox{90}{\makebox[\tempdim]{\small (a) Input }}\hfil
    \begin{minipage}[b]{.95\linewidth}
        \includegraphics[width=.24\linewidth]{figs/limitation/input/0002.jpg}
        \includegraphics[width=.24\linewidth]{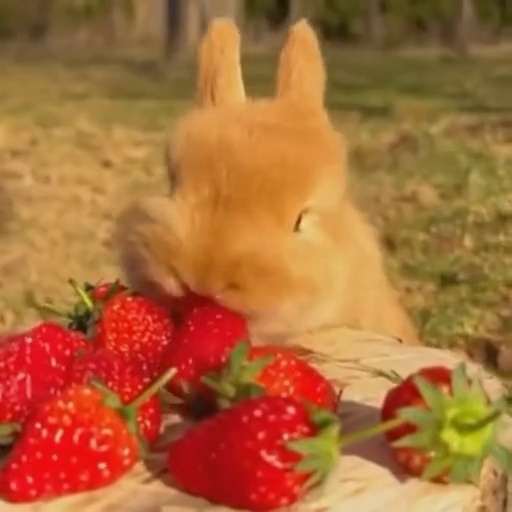}
        \includegraphics[width=.24\linewidth]{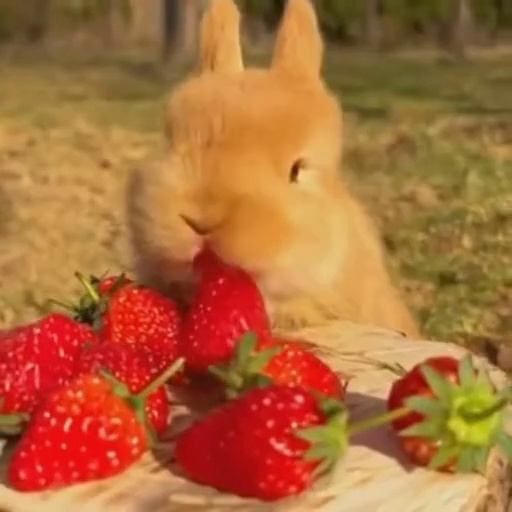}
        \includegraphics[width=.24\linewidth]{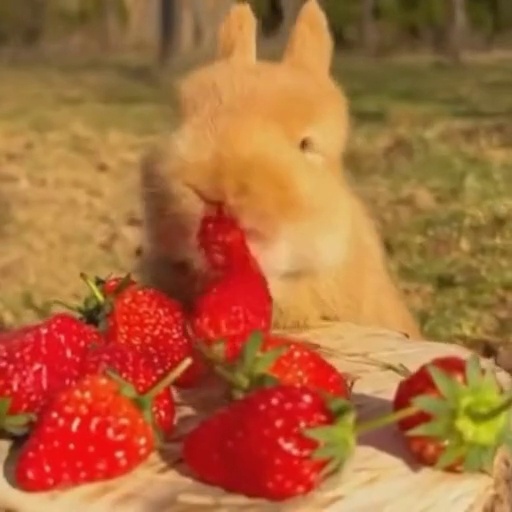}
    \end{minipage}
    \rotatebox{90}{\makebox[\tempdim]{\small (b) Ours }}\hfil
    \begin{minipage}[b]{.95\linewidth}
        \includegraphics[width=.24\linewidth]{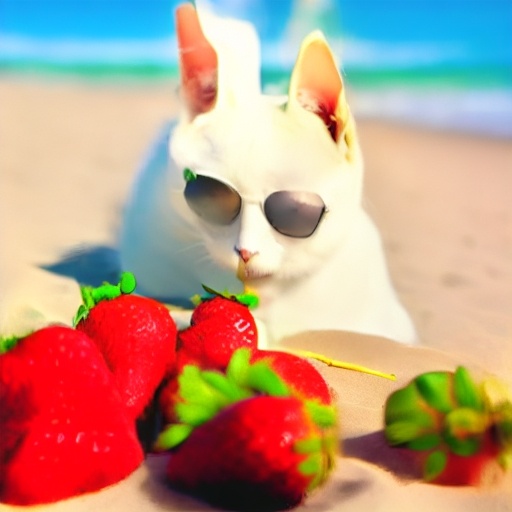}
        \includegraphics[width=.24\linewidth]{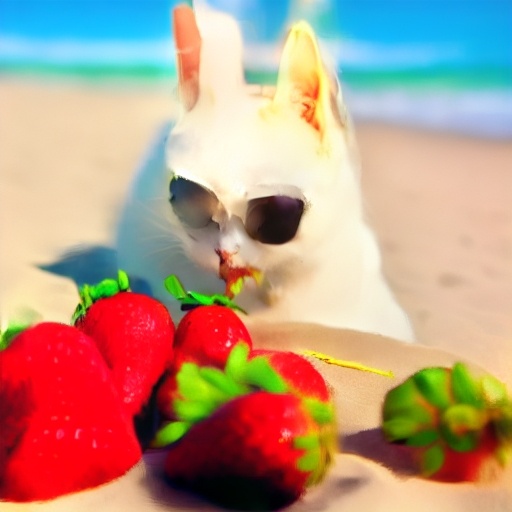}
        \includegraphics[width=.24\linewidth]{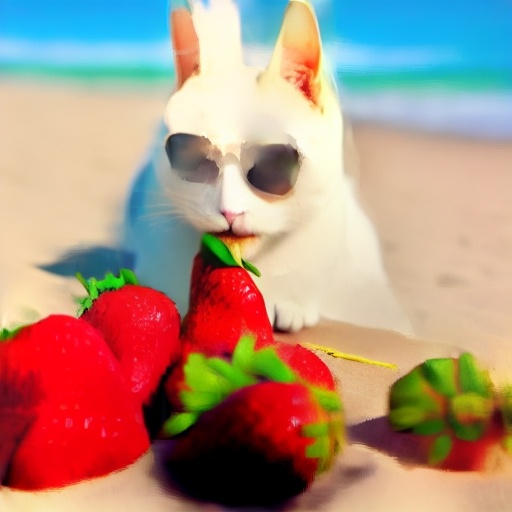}
        \includegraphics[width=.24\linewidth]{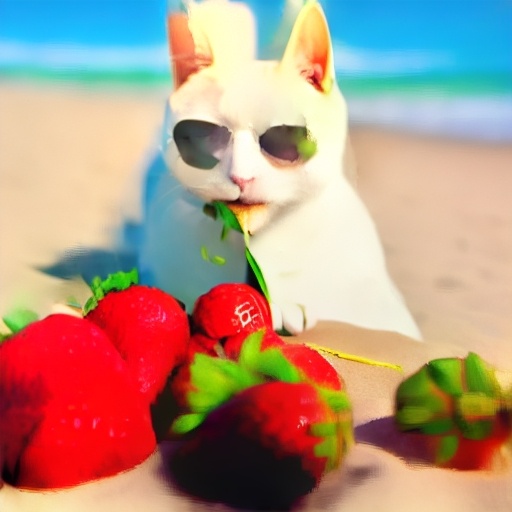}
    \end{minipage}
    \vspace{-.05in}
    \caption{\textbf{Failure case.} Inconsistent stylized regions or undesirable 
    deformation (e.g. sunglasses) may be resulted with inaccurate optical flow. Text prompt: {\footnotesize\tt "A cat with sunglasses is eating a strawberry on the beach."}}
    \vspace{-.05in}
    \label{fig:limitation}
\end{figure}

Firstly, our multi-frame fusion steps rely on optical flow for information sharing. Therefore, inaccurate optical flow estimation may lead to inconsistent appearance. 
Moreover, our proposed method may fail to change the geometry of the original video as we rely on the Canny edge condition. In Fig.~\ref{fig:limitation}, when changing the rabbit to a cat, the optical flow at the area with geometry changes will be incorrect, resulting in distortion and blurriness at the ears and sunglasses.

\section{Conclusion}
We propose a zero-shot text-driven approach for video stylization. We design a multi-frame fusion module to generate stylized videos with high-detailed fidelity and temporal consistency. We utilize the optical flow of the original video as a correspondence site to share information among edited frames. 
Our extensive experiments and demonstrate that our approach achieves outstanding qualitative and quantitative results compared to state-of-the-art methods. Unlike the previous methods which may  exhibit serious visual artifacts of certain forms, our method produce high-quality results that highly respects the user text prompt semantically, and simultaneously,respects the motion in the given video.

{
    \small
    \bibliographystyle{ieeenat_fullname}
    \bibliography{main}
}

% WARNING: do not forget to delete the supplementary pages from your submission 
% \input{sec/X_suppl}

\end{document}